%% file: conference_101719.tex
\documentclass[conference]{IEEEtran}
\IEEEoverridecommandlockouts
\usepackage{cite}
\usepackage{amsmath,amssymb,amsfonts}
\usepackage{algorithmic}
\usepackage{graphicx}
\usepackage{textcomp}
\usepackage{xcolor}

\usepackage{amsmath,amssymb,amsfonts}
\usepackage{algorithmic}
\usepackage{subcaption}
\usepackage{graphicx}
\usepackage{textcomp}
\usepackage{xcolor}
\usepackage{paralist}
\usepackage{pgf}
\usepackage{tikz}
\usepackage{booktabs}
\usepackage{tabularx}
\usepackage{hyperref}
\usepackage{titlesec}
\usepackage{orcidlink}

\usepackage{fancyhdr}

\fancypagestyle{firstpage}{
  \fancyhf{}

  \fancyhead[C]{2026 IEEE 33rd International Conference on Electronics, Circuits and Systems (ICECS)}
  \fancyfoot[L]{979-8-3195-1905-4/26/\$31.00~\copyright~2026~IEEE}
}

\usepackage{pgfplots}
\pgfplotsset{compat=newest}
\usetikzlibrary{patterns}
\usepgfplotslibrary{fillbetween}
\def\BibTeX{{\rm B\kern-.05em{\sc i\kern-.025em b}\kern-.08em
    T\kern-.1667em\lower.7ex\hbox{E}\kern-.125emX}}

    \DeclareMathOperator*{\argmin}{argmin}
\newcommand*\circled[1]{\tikz[baseline=(char.base)]{
            \node[shape=circle,draw,inner sep=0.8pt] (char) {#1};}}
\begin{document}

\title{Carbon-Aware Routing for Function Calling in Edge-Cloud LLM Systems
}

\author{
\IEEEauthorblockN{
Aikaterini Maria Panteleaki\,\orcidlink{0009-0005-7827-8875},
Varatheepan Paramanayakam\,\orcidlink{0009-0000-1982-9693},
Spyros Tragoudas\,\orcidlink{0009-0006-2575-3588},
Iraklis Anagnostopoulos\,\orcidlink{0000-0003-0985-3045}
}

\IEEEauthorblockA{
\textit{School of Electrical, Computer and Biomedical Engineering} \\
\textit{Southern Illinois University Carbondale} \\
aikaterinimaria.panteleaki@siu.edu,
varatheepan@siu.edu,
spyros@siu.edu,
iraklis.anagno@siu.edu
}
}

\maketitle
\thispagestyle{firstpage}
\begin{abstract}
Large Language Models (LLMs) with function-calling capabilities are becoming critical for modern agentic AI systems. Nevertheless, current deployments typically route inferences to powerful cloud-based models, incurring significant energy use and carbon emissions. We address this sustainability challenge with a carbon-aware routing framework that distributes function-calling queries across a three-tier edge-cloud architecture, combining edge and cloud LLMs on heterogeneous hardware. At its core, a lightweight k-NN predictor operating in a unified semantic-lexical embedding space estimates query-specific accuracy, delay, and power consumption on each edge tier. These predictions are then combined with real-time grid carbon intensity to route every query to the lowest-emission tier capable of executing it successfully. Evaluated on state-of-the-art function-calling benchmarks and LLM families, our framework matches cloud-level accuracy while reducing operational carbon emissions by $4\times$ on average.
\end{abstract}


\input{introduction}
\input{methodology}
\input{evaluation}
\input{conclusion}
\input{acknowledgement}

\bibliographystyle{IEEEtran}
\bibliography{ref}

\end{document}

%% file: introduction.tex
\section{Introduction}

Large Language Models (LLMs) have evolved from text generation systems to sophisticated decision-making agents with interactive tool-use capabilities. The key capability enabling this evolution is \textit{\textbf{function calling}}, the process by which an agent autonomously selects and invokes specific functions from available tools enabling it to: (i) access external data sources, (ii) execute computations beyond the LLM's parametric knowledge, and (iii)  interact with software systems and APIs. Function calling serves as a fundamental building block of modern agentic AI systems \cite{patil2025bfcl}, but its complexity and accuracy requirements push deployments toward an expensive default of routing inferences to powerful cloud-based models, regardless of how trivial the task itself is~\cite{erdogan-etal-2024-tinyagent}. Prior work on sustainability-aware LLM execution at the edge has shown that dynamic tool selection and carbon-aware power adaptation can reduce emissions by up to 52\% on edge hardware~\cite{paramanayakam2025carboncall}.

The environmental impact of this approach is substantial. ChatGPT alone serves over 2.5 billion queries per day, consuming approximately 850 megawatt-hours, enough to fully charge 14,000 electric vehicles~\cite{smith2025hidden}. Annually, ChatGPT's energy consumption reaches 310 gigawatt-hours, roughly equivalent to powering 29,000 U.S. homes for a whole year. Including competitors like Gemini and Claude, the combined annual power draw reaches 15 terawatt-hours, similar to the output of two nuclear reactors \cite{smith2025hidden}. Crucially, inference is now the dominant contributor to this footprint, as reports from cloud providers indicate that inference accounts for 80-90\% of total Machine Learning (ML) cloud compute demand and 60\% of ML-related energy consumption \cite{luccioni2024power}. As function-calling agents continue to grow, the always-cloud default will only amplify this trend.

\begin{table}[h]
\centering
\vspace{-0.8em}
\caption{Success Rate on BFCL V2~\cite{patil2025bfcl}, Energy Consumption per query and estimated Carbon Emissions for LLMs}
\vspace{-0.8em}
\resizebox{\columnwidth}{!}{%
\begin{tabular}{lccc}
\toprule
\textbf{Model} & \textbf{Success Rate} & \textbf{General Energy/Query} & \textbf{Carbon Emissions}~\cite{powerbiDashboard2025}  \\
\midrule
Llama3.1-70B (Cloud) & 97.8\% & 4.35 Wh~\cite{powerbiDashboard2025} & $\sim$ 1.30 $\mathrm{gCO_2eq}$  \\
Llama3.1-8B (Edge) & 56.5\% & 0.17 Wh~\cite{jang2025edge} & $\sim$ 0.05 $\mathrm{gCO_2eq}$  \\
Llama3.2-1B (Edge) & 13.6\% & 0.04 Wh~\cite{jang2025edge} & $\sim$ 0.01 $\mathrm{gCO_2eq}$  \\
\bottomrule
\end{tabular}%
}
\vspace{-5pt}
\label{tab:accuracy_energy_tradeoff}
\end{table}

Edge-based solutions offer a promising alternative. As shown in Table \ref{tab:accuracy_energy_tradeoff}, edge-deployed models like Llama3.1-8B consume approximately 25$\times$ less energy per query than their 70B cloud counterparts, with proportional savings in carbon emissions \cite{jang2025edge, powerbiDashboard2025}. Accuracy, however, remains the bottleneck. On Berkeley Function-Calling Leaderboard V2 (BFCL V2)~\cite{patil2025bfcl} Llama3.1-70B achieves 97.8\% success rate, while the smaller edge variants Llama3.1-8B and Llama3.2-1B reach only 56.5\% and 13.6\% respectively.
This accuracy gap, however, is not uniform across queries. Many function-calling tasks, such as single-tool invocations, simple retrieval, or structured queries, do not require the reasoning capabilities of large-scale models and can be served correctly by small edge models. This inherent heterogeneity creates a design opportunity: \emph{Can we dynamically allocate function-calling queries across heterogeneous edge–cloud tiers to minimize carbon footprint while maintaining task-level correctness?} While carbon-aware and power-efficient execution has been explored for DNN workloads through attention-based multi-DNN management and carbon-driven power capping on edge servers~\cite{karatzas2024mapformer,paramanayakam2025ecomap}, extending such principles to LLM function calling remains an open challenge.

Existing LLM routers \cite{stripelis2024tensoropera,ding2024hybrid,paramanayakam2025less} dispatch queries across heterogeneous models to balance accuracy against dollar cost, but ignore the underlying hardware, power and carbon intensity of the electricity grid powering each tier. Without these signals, even routers that successfully shift load off the cloud capture only part of the available carbon savings, since they cannot identify the greenest tier for a given query at a given moment.
In this paper, we treat function-calling inference as a Performance–Power–Carbon (PPC) constrained routing problem and present a carbon-aware framework that solves it. Queries are distributed across a three-tier edge–cloud architecture including tiny/micro LLMs (1–4B) on a low-power edge SoC, small LLMs (7–12B) on an edge accelerator, and large LLMs ($>$100B) hosted in the cloud. The framework is developed within NSF project 2324854 on sustainable AI~\cite{panteleaki2024carbon,panteleaki2025vertical}, with main contributions:
\begin{inparaenum}
    \item[\circled{1}] \textbf{Analytic Performance Modeling.} We develop a lightweight k-NN regression model operating in a unified semantic-lexical embedding space that acts as an analytic predictor of per-tier probability of correct execution, inference delay, and carbon footprint. By learning from historical executions, this model enables rapid design-space exploration without invoking the underlying LLMs.
    \item[\circled{2}] \textbf{Carbon-Aware Routing Policy.} Using these predictions, we formulate a routing strategy that selects the most carbon-efficient tier satisfying task-level accuracy constraints, while adapting to real-time grid carbon intensity.
    \item[\circled{3}] \textbf{Comprehensive Evaluation.} Through experiments on state-of-the-art function-calling benchmarks, we show that our approach preserves cloud-level success rates while reducing carbon emissions by an average of $4\times$ and up to $8\times$ for simpler queries, demonstrating the effectiveness of PPC-aware tiered inference.
\end{inparaenum}

%% file: methodology.tex
\section{Methodology}

\begin{figure}[t]
       \centering
       \resizebox{1\columnwidth}{!}{\includegraphics{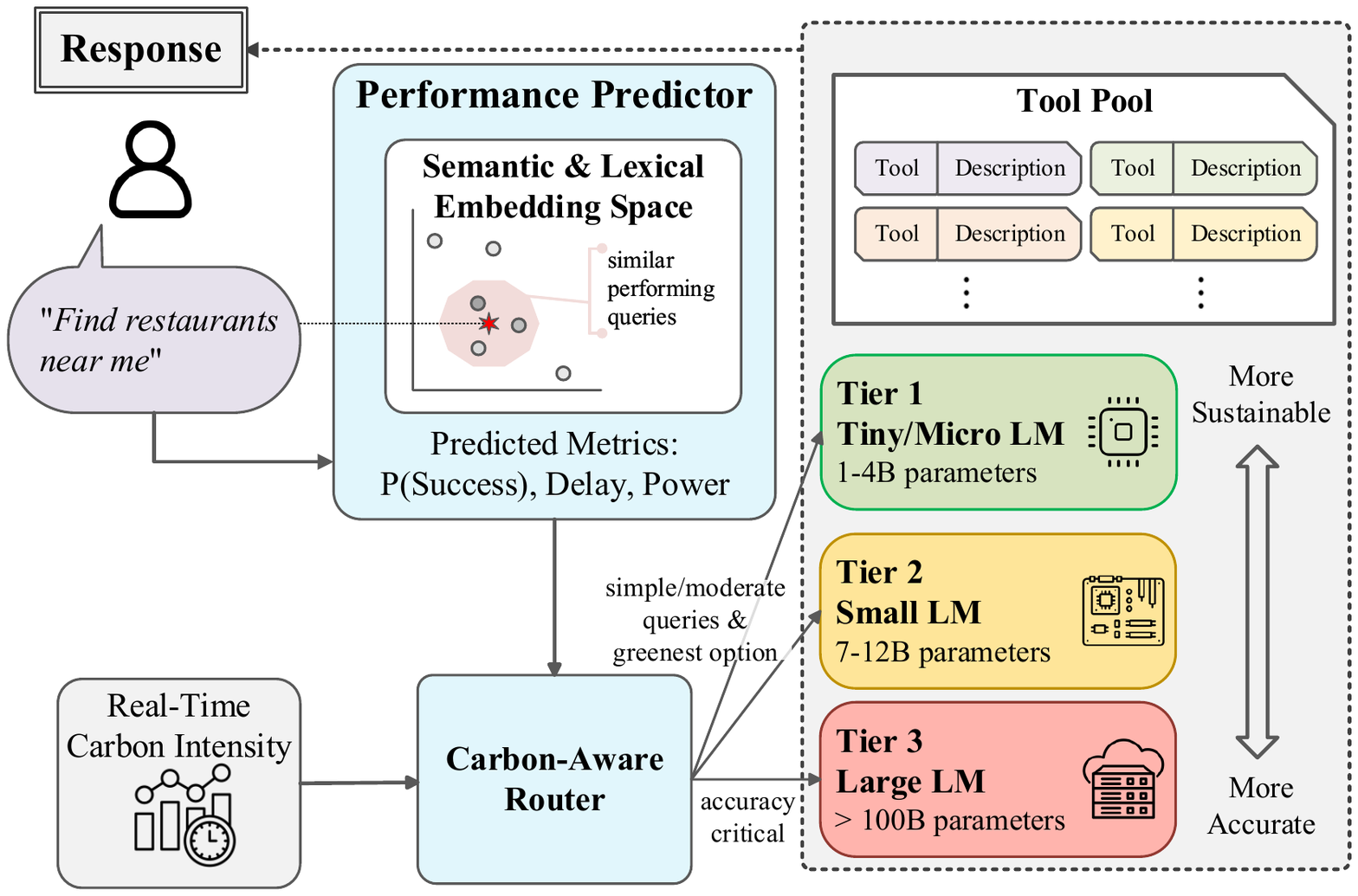}}
       \caption{Carbon-aware routing framework architecture}
       \label{fig:overview}
\end{figure}

\noindent\textbf{\textit{System Architecture}}
Fig.\ref{fig:overview} provides an overview of our framework. We consider a three-tier heterogeneous edge–cloud architecture, where each tier represents a distinct combination of model capability and hardware platform, providing different trade-offs between accuracy and carbon efficiency:
\begin{inparaenum}
    \item[\circled{1}] \textit{Tier~1} (Tiny/Micro LM) deploys models with 1-4B parameters on resource-constrained edge devices, consuming the lowest energy per query, but offering limited reasoning capabilities for complex tasks,
    \item[\circled{2}] \textit{Tier~2} (Small LM) hosts models with 7-12B parameters on edge infrastructure, offering higher accuracy than Tier~1 with energy consumption well below cloud levels, and
    \item[\circled{3}] \textit{Tier~3} (Large LM) runs cloud-based models with more than 100B parameters deployed on datacenter infrastructure, delivering state-of-the-art accuracy at  the highest environmental cost.
\end{inparaenum}

At runtime, tier selection is performed by 
\begin{inparaenum}[(i)]
	\item a performance predictor and
	\item a carbon-aware router
\end{inparaenum}, both shown in Fig.\ref{fig:overview}. The Performance Predictor estimates, for each query-tier pair on the edge, the probability of correct execution, the inference delay, and the power consumption. The Carbon-Aware Router combines these predictions with the real-time carbon intensity of each tier's grid and assigns every query to the lowest-emission tier whose predicted success probability satisfies a configurable accuracy threshold, falling back to the cloud only when no edge tier qualifies.

\noindent\textbf{\textit{Performance Prediction}}
A key requirement for carbon-aware routing is the ability to estimate, prior to execution, how a query will behave on each edge tier. For each pair of query $q_i$ and edge tier $\tau \in \{\text{Tier~1}, \text{Tier~2}\}$, our performance predictor outputs the probability of correct execution $P({\text{Success}}(q_i, \tau))$, the inference delay ${\text{D}}(q_i, \tau)$, and the power consumption ${\text{w}}(q_i, \tau)$. Together, these three quantities determine the expected carbon footprint of routing the query to that tier.

The predictor operates through an offline calibration phase, where each calibration query is encoded into a semantic-lexical embedding space. This space combines semantic embeddings, capturing the linguistic and structural content of the query, along with lexical complexity scores that measure low-frequency word usage, a signal shown to correlate with how LLMs perceive query difficulty~\cite{kelious-etal-2025-large}. For each calibration query, we record the binary Success outcome against ground truth, together with the measured inference delay and power on the target hardware. Unlike other generative AI tasks, function calling does not tolerate partial correctness, since invoking the wrong tool or missing a required one causes the entire response to fail, and the binary Success metric reflects this.

At runtime, each incoming query is mapped into the embedding space. A k-Nearest Neighbor (k-NN) regression model identifies the $k$ most similar calibration queries using cosine distance, and produces distance-weighted predictions of success probability, delay and power for each edge tier. Using the same neighborhood for all three quantities ensures that the predictions remain mutually consistent, capturing joint correlations such as the tendency of complex queries to exhibit both reduced accuracy and increased latency on tiny models~\cite{xu2019survey}. Unlike ML routing models, which require training and risk overfitting, our k-NN is non-parametric, requires no training and incurs negligible runtime overhead.

\noindent\textbf{\textit{Carbon-Aware Routing}}
To compute the carbon cost of each routing decision, we combine the predicted delay and power with the real-time carbon intensity of each tier's grid. The carbon footprint of executing query $q_i$ on tier $\tau$ is:
\begin{equation}
    \text{CF}(q_i, \tau) = \text{CI}_\tau  \cdot \text{D}(q_i, \tau)  \cdot  \text{w}(q_i, \tau)
    \label{eq:carbon}
\end{equation}
where $\text{CI}_\tau$ is the real-time carbon intensity of the grid powering tier $\tau$, measured in $\mathrm{gCO_2}/\mathrm{kWh}$, while $\text{D}(q_i, \tau)$ and $\text{w}(q_i, \tau)$ are the predicted inference delay and power consumption obtained from the performance predictor.

The routing policy dynamically assigns each query to the most carbon-efficient tier capable of successful execution. We first identify the edge tiers whose predicted Success probability satisfies the minimum accuracy threshold, $P({\text{Success}}(q_i, \tau)) \geq \theta$, and among them select the one with the minimum estimated carbon footprint. If no edge tier qualifies, the query is routed to Tier~3:

\begin{equation}
    \tau_i^* = 
    \begin{cases}
        \displaystyle\argmin_{\tau \in \{\text{Tier 1, Tier 2}\}} \text{CF}(q_i, \tau), & \text{if } \exists\, \ \tau: P(\text{Success}) \geq \theta \\
        \text{Tier 3}, & \text{otherwise}
    \end{cases}
    \label{eq:opt}
\end{equation}

This policy ensures that Tier~3 is triggered only when no edge tier meets the accuracy requirement, minimizing the use of the highest-emission tier.

%% file: evaluation.tex
\section{Evaluation}

\noindent\textbf{\textit{Experimental Setup}}
We evaluate our framework on two function-calling benchmarks with distinct characteristics: BFCL V2~\cite{patil2025bfcl} and GeoEngine~\cite{singh2024geoengine}. BFCL V2 covers tasks from diverse domains such as mathematics, coding, and general reasoning, while GeoEngine targets geospatial applications with elevated difficulty, requiring sequential function calls where each depends on the previous result. For each benchmark, we employ an 80-20 split, using 20\% of the queries for calibration and 80\% for evaluation. We report Success Rate as the percentage of successfully executed queries against ground truth, and set the accuracy threshold to $\theta$ = 0.9.

Real-time carbon intensity values for each tier's geographic location are provided by the Electricity Maps API~\cite{electricitymaps2025}. Tier~1 is deployed on an NVIDIA Jetson Orin Nano~\cite{nvidia_orin_nano_datasheet}, Tier~2 on the more capable NVIDIA Jetson AGX Orin~\cite{NVIDIA_AGX_Orin}, and Tier~3 hosts a cloud LLM on H100 SXM GPUs through the Fireworks AI platform~\cite{FireworksAI}.
Tier~3 runs GPT-OSS-120B from OpenAI across all configurations. For Tiers~1 and~2, we evaluate our routing policy across four model families to demonstrate generalization: Qwen3 (1.7B, 8B), Gemma3 (4B, 12B), Llama3.1 (xLAM-1B, 8B), and Falcon3 (3B, 7B).

\noindent\textbf{\textit{Baseline Routing Methods}}
We compare our carbon-aware routing framework against five baselines.
\begin{inparaenum}
    \item[\circled{1}] \textit{Cloud~Only} forwards all queries to Tier~3, maximizing accuracy at the highest carbon cost. 
    \item[\circled{2}] \textit{Edge~Only} routes all queries to Tier~2, prioritizing local execution for carbon reduction, at the cost of accuracy.
    \item[\circled{3}]  \textit{TensorOpera} \cite{stripelis2024tensoropera} implements a multi-model router optimizing for accuracy and cost. It serves as our carbon-unaware counterpart, to quantify the gains from carbon awareness. 
    \item[\circled{4}] \textit{HybridLLM} \cite{ding2024hybrid} uses a learned two-way router to assign each query to either a small or large model. We adapt it to route between Tier 2 and 3, isolating the carbon savings of additionally exploiting Tier~1.
    \item[\circled{5}]  \textit{Less-is-More} \cite{paramanayakam2025less} optimizes function-calling efficiency on edge devices by dynamically reducing the available tool set, to match query complexity. We restrict its deployment to Tier 2, since it was designed as a single-device optimization rather than a multi-tier router.
\end{inparaenum}

\begin{figure}
  \vspace{-0.5em}
      \centering
      \input{figures/shared_legend.pgf}
      \begin{subfigure}{\columnwidth}
          \centering
          \footnotesize
          \vspace{-0.9em}
          \input{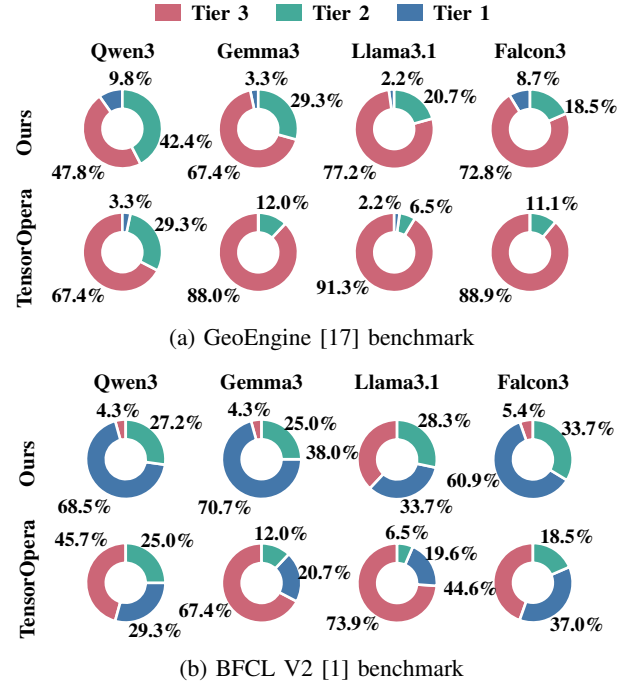}
          \vspace{-0.8em}
          \caption{GeoEngine\cite{singh2024geoengine} benchmark}
          \label{fig:queries_geoengine}
      \end{subfigure}
      
      \begin{subfigure}{\columnwidth}
          \centering
          \vspace{0.1em}
          \footnotesize
          \input{figures/gorilla_distribution_new.pgf}
          \vspace{-0.8em}
          \caption{BFCL V2 \cite{patil2025bfcl} benchmark}
          \label{fig:queries_gorilla}
      \end{subfigure}
      \caption{Query Distribution across three-tier architecture comparing our framework and TensorOpera~\cite{stripelis2024tensoropera}}
      \label{queries}
      \vspace{-1em}
  \end{figure}

 \begin{figure*}[t]
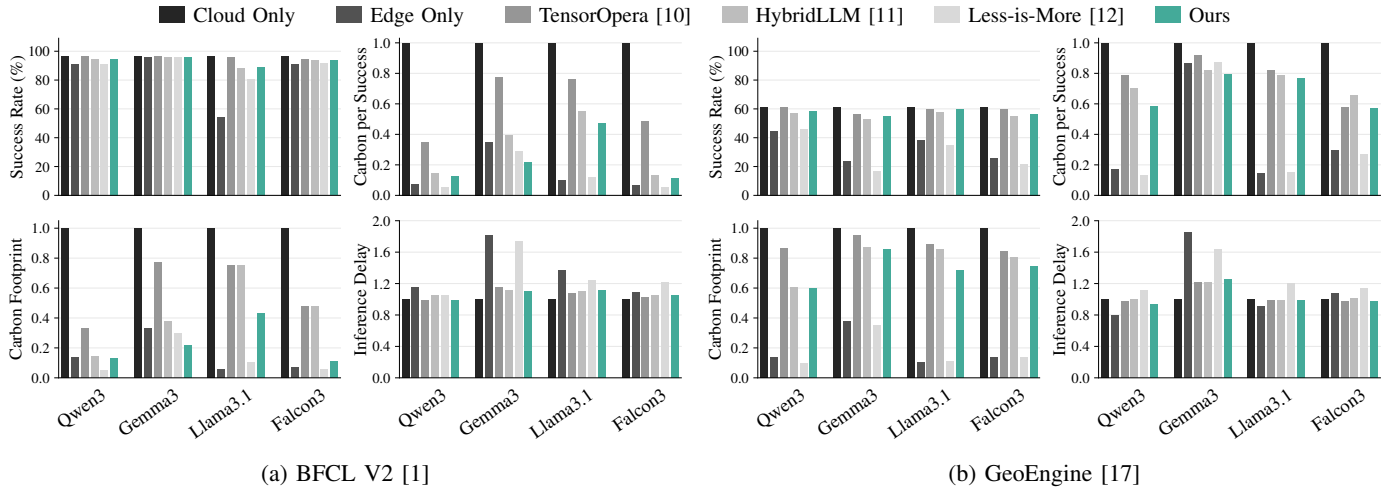

  \centering
  \input{figures/paper_legend.pgf}
  \begin{subfigure}{0.49\linewidth}
      \centering
      \resizebox{\linewidth}{!}{\input{figures/gorilla_paper_combined.pgf}}
      \caption{BFCL V2~\cite{patil2025bfcl}}
  \end{subfigure}\hfill
  \begin{subfigure}{0.49\linewidth}
      \centering
      \resizebox{\linewidth}{!}{\input{figures/geoengine_paper_combined.pgf}}
      \caption{GeoEngine~\cite{singh2024geoengine}}
  \end{subfigure}

  \caption{Performance comparison on BFCL V2 and GeoEngine: Success Rate (\%), Normalized Carbon Footprint, Carbon-per-Success, and Normalized Inference Delay. All normalized metrics use Cloud~Only as the reference.}
  \label{fig:results_combined}
  \end{figure*}

\noindent\textbf{\textit{Results Discussion}}
Figure \ref{queries} shows how each method splits queries across tiers for both benchmarks. For GeoEngine, which requires complex sequential function calling, our carbon-aware framework distributes queries across all three tiers. For instance, with Qwen3 our method routes 42.4\% to Tier~3, 47.8\% to Tier~2 and 9.8\% to Tier~1, exploiting edge resources even for this challenging dataset. The Llama3.1 configuration is more cloud-heavy at 77.2\%, but still executes 22.8\% on the edge. In contrast, TensorOpera routes the vast majority of queries to Tier~3, ranging from 67.4\% for Qwen3 to 91.3\% for Llama3.1 prioritizing accuracy at increased carbon cost.
On the simpler BFCL V2 benchmark, differences become more prominent. Our framework shifts aggressively to the edge, with Qwen3 and Gemma3 routing almost 70\% to Tier~1 and only 4.3\% to Tier~3, while TensorOpera continues to send 67.4\% - 73.9\% to the cloud, despite the lower query complexity. This contrast demonstrates our router's ability to dynamically balance accuracy and carbon efficiency according to query complexity, whereas TensorOpera's accuracy-cost objective favors cloud, regardless of need.

Figure \ref{fig:results_combined} presents success rate, carbon, and delay metrics across all model families for both datasets. On GeoEngine, our framework achieves 55.3-59.6\% success across families, closely matching the cloud-only baseline at 60.9\%, while reducing carbon footprint by approximately $1.39\times$ on average. Edge-only achieves the lowest emissions but unacceptably low success rates, confirming that edge-only execution fails for complex function calling. TensorOpera and HybridLLM achieve competitive success rates but with elevated carbon cost due to their cloud-intensive routing, while Less-is-More achieves low success rates. On BFCL V2, our framework achieves 89.1-95.7\% success rates with negligible accuracy loss compared to cloud-only's 96.7\%, while reaching $6\times$ carbon reduction on average and up to $8\times$ for Falcon3. The carbon-per-success efficiency metric confirms that these gains are not at the expense of correctness, as our framework spends less carbon to produce a correct answer, than any baseline at comparable accuracy. In contrast, edge-focused methods like Less-is-More show notably low absolute carbon but worse efficiency because of their lower success rate. Edge-tier delay efficiency also varies by model family. Gemma3, for instance, is consistently the slowest on our edge hardware across both benchmarks. Despite this, our router keeps overall inference delay close to cloud-only levels, confirming that the carbon savings do not come at the expense of responsiveness.

%% file: figures/shared_legend.pgf
\begingroup%
\makeatletter%
\begin{pgfpicture}%
\pgfpathrectangle{\pgfpointorigin}{\pgfqpoint{1.932770in}{0.303721in}}%
\pgfusepath{use as bounding box, clip}%
\begin{pgfscope}%
\pgfsetbuttcap%
\pgfsetmiterjoin%
\definecolor{currentfill}{rgb}{1.000000,1.000000,1.000000}%
\pgfsetfillcolor{currentfill}%
\pgfsetlinewidth{0.000000pt}%
\definecolor{currentstroke}{rgb}{1.000000,1.000000,1.000000}%
\pgfsetstrokecolor{currentstroke}%
\pgfsetdash{}{0pt}%
\pgfpathmoveto{\pgfqpoint{0.000000in}{-0.000000in}}%
\pgfpathlineto{\pgfqpoint{1.932770in}{-0.000000in}}%
\pgfpathlineto{\pgfqpoint{1.932770in}{0.303721in}}%
\pgfpathlineto{\pgfqpoint{0.000000in}{0.303721in}}%
\pgfpathlineto{\pgfqpoint{0.000000in}{-0.000000in}}%
\pgfpathclose%
\pgfusepath{fill}%
\end{pgfscope}%
\begin{pgfscope}%
\pgfsetbuttcap%
\pgfsetmiterjoin%
\definecolor{currentfill}{rgb}{0.800000,0.400000,0.466667}%
\pgfsetfillcolor{currentfill}%
\pgfsetlinewidth{0.000000pt}%
\definecolor{currentstroke}{rgb}{0.000000,0.000000,0.000000}%
\pgfsetstrokecolor{currentstroke}%
\pgfsetstrokeopacity{0.000000}%
\pgfsetdash{}{0pt}%
\pgfpathmoveto{\pgfqpoint{0.100000in}{0.118721in}}%
\pgfpathlineto{\pgfqpoint{0.233333in}{0.118721in}}%
\pgfpathlineto{\pgfqpoint{0.233333in}{0.203721in}}%
\pgfpathlineto{\pgfqpoint{0.100000in}{0.203721in}}%
\pgfpathlineto{\pgfqpoint{0.100000in}{0.118721in}}%
\pgfpathclose%
\pgfusepath{fill}%
\end{pgfscope}%
\begin{pgfscope}%
\definecolor{textcolor}{rgb}{0.000000,0.000000,0.000000}%
\pgfsetstrokecolor{textcolor}%
\pgfsetfillcolor{textcolor}%
\pgftext[x=0.288889in,y=0.122610in,left,base]{\color{textcolor}{\rmfamily\fontsize{8.000000}{9.600000}\bfseries\selectfont\catcode`\^=\active\def^{\ifmmode\sp\else\^{}\fi}\catcode`\%=\active\def
\end{pgfscope}%
\begin{pgfscope}%
\pgfsetbuttcap%
\pgfsetmiterjoin%
\definecolor{currentfill}{rgb}{0.266667,0.666667,0.600000}%
\pgfsetfillcolor{currentfill}%
\pgfsetlinewidth{0.000000pt}%
\definecolor{currentstroke}{rgb}{0.000000,0.000000,0.000000}%
\pgfsetstrokecolor{currentstroke}%
\pgfsetstrokeopacity{0.000000}%
\pgfsetdash{}{0pt}%
\pgfpathmoveto{\pgfqpoint{0.729442in}{0.118721in}}%
\pgfpathlineto{\pgfqpoint{0.862775in}{0.118721in}}%
\pgfpathlineto{\pgfqpoint{0.862775in}{0.203721in}}%
\pgfpathlineto{\pgfqpoint{0.729442in}{0.203721in}}%
\pgfpathlineto{\pgfqpoint{0.729442in}{0.118721in}}%
\pgfpathclose%
\pgfusepath{fill}%
\end{pgfscope}%
\begin{pgfscope}%
\definecolor{textcolor}{rgb}{0.000000,0.000000,0.000000}%
\pgfsetstrokecolor{textcolor}%
\pgfsetfillcolor{textcolor}%
\pgftext[x=0.918331in,y=0.122610in,left,base]{\color{textcolor}{\rmfamily\fontsize{8.000000}{9.600000}\bfseries\selectfont\catcode`\^=\active\def^{\ifmmode\sp\else\^{}\fi}\catcode`\%=\active\def
\end{pgfscope}%
\begin{pgfscope}%
\pgfsetbuttcap%
\pgfsetmiterjoin%
\definecolor{currentfill}{rgb}{0.266667,0.466667,0.666667}%
\pgfsetfillcolor{currentfill}%
\pgfsetlinewidth{0.000000pt}%
\definecolor{currentstroke}{rgb}{0.000000,0.000000,0.000000}%
\pgfsetstrokecolor{currentstroke}%
\pgfsetstrokeopacity{0.000000}%
\pgfsetdash{}{0pt}%
\pgfpathmoveto{\pgfqpoint{1.358884in}{0.118721in}}%
\pgfpathlineto{\pgfqpoint{1.492217in}{0.118721in}}%
\pgfpathlineto{\pgfqpoint{1.492217in}{0.203721in}}%
\pgfpathlineto{\pgfqpoint{1.358884in}{0.203721in}}%
\pgfpathlineto{\pgfqpoint{1.358884in}{0.118721in}}%
\pgfpathclose%
\pgfusepath{fill}%
\end{pgfscope}%
\begin{pgfscope}%
\definecolor{textcolor}{rgb}{0.000000,0.000000,0.000000}%
\pgfsetstrokecolor{textcolor}%
\pgfsetfillcolor{textcolor}%
\pgftext[x=1.547773in,y=0.122610in,left,base]{\color{textcolor}{\rmfamily\fontsize{8.000000}{9.600000}\bfseries\selectfont\catcode`\^=\active\def^{\ifmmode\sp\else\^{}\fi}\catcode`\%=\active\def
\end{pgfscope}%
\end{pgfpicture}%
\makeatother%
\endgroup%

%% file: figures/gorilla_distribution_new.pgf
\begingroup%
\makeatletter%
\begin{pgfpicture}%
\pgfpathrectangle{\pgfpointorigin}{\pgfqpoint{3.307239in}{1.576217in}}%
\pgfusepath{use as bounding box, clip}%
\begin{pgfscope}%
\pgfsetbuttcap%
\pgfsetmiterjoin%
\definecolor{currentfill}{rgb}{1.000000,1.000000,1.000000}%
\pgfsetfillcolor{currentfill}%
\pgfsetlinewidth{0.000000pt}%
\definecolor{currentstroke}{rgb}{1.000000,1.000000,1.000000}%
\pgfsetstrokecolor{currentstroke}%
\pgfsetdash{}{0pt}%
\pgfpathmoveto{\pgfqpoint{0.000000in}{0.000000in}}%
\pgfpathlineto{\pgfqpoint{3.307239in}{0.000000in}}%
\pgfpathlineto{\pgfqpoint{3.307239in}{1.576217in}}%
\pgfpathlineto{\pgfqpoint{0.000000in}{1.576217in}}%
\pgfpathlineto{\pgfqpoint{0.000000in}{0.000000in}}%
\pgfpathclose%
\pgfusepath{fill}%
\end{pgfscope}%
\begin{pgfscope}%
\pgfsetbuttcap%
\pgfsetmiterjoin%
\definecolor{currentfill}{rgb}{0.800000,0.400000,0.466667}%
\pgfsetfillcolor{currentfill}%
\pgfsetlinewidth{1.007514pt}%
\definecolor{currentstroke}{rgb}{1.000000,1.000000,1.000000}%
\pgfsetstrokecolor{currentstroke}%
\pgfsetdash{}{0pt}%
\pgfpathmoveto{\pgfqpoint{0.652372in}{1.242797in}}%
\pgfpathcurveto{\pgfqpoint{0.643045in}{1.242797in}}{\pgfqpoint{0.633729in}{1.242167in}}{\pgfqpoint{0.624488in}{1.240911in}}%
\pgfpathcurveto{\pgfqpoint{0.615246in}{1.239655in}}{\pgfqpoint{0.606100in}{1.237776in}}{\pgfqpoint{0.597111in}{1.235286in}}%
\pgfpathlineto{\pgfqpoint{0.624742in}{1.135519in}}%
\pgfpathcurveto{\pgfqpoint{0.629236in}{1.136764in}}{\pgfqpoint{0.633809in}{1.137703in}}{\pgfqpoint{0.638430in}{1.138332in}}%
\pgfpathcurveto{\pgfqpoint{0.643051in}{1.138960in}}{\pgfqpoint{0.647709in}{1.139275in}}{\pgfqpoint{0.652372in}{1.139275in}}%
\pgfpathlineto{\pgfqpoint{0.652372in}{1.242797in}}%
\pgfpathclose%
\pgfusepath{stroke,fill}%
\end{pgfscope}%
\begin{pgfscope}%
\pgfsetbuttcap%
\pgfsetmiterjoin%
\definecolor{currentfill}{rgb}{0.266667,0.466667,0.666667}%
\pgfsetfillcolor{currentfill}%
\pgfsetlinewidth{1.007514pt}%
\definecolor{currentstroke}{rgb}{1.000000,1.000000,1.000000}%
\pgfsetstrokecolor{currentstroke}%
\pgfsetdash{}{0pt}%
\pgfpathmoveto{\pgfqpoint{0.597111in}{1.235286in}}%
\pgfpathcurveto{\pgfqpoint{0.561114in}{1.225317in}}{\pgfqpoint{0.528476in}{1.205790in}}{\pgfqpoint{0.502673in}{1.178783in}}%
\pgfpathcurveto{\pgfqpoint{0.476869in}{1.151776in}}{\pgfqpoint{0.458848in}{1.118283in}}{\pgfqpoint{0.450528in}{1.081869in}}%
\pgfpathcurveto{\pgfqpoint{0.442208in}{1.045455in}}{\pgfqpoint{0.443896in}{1.007460in}}{\pgfqpoint{0.455410in}{0.971926in}}%
\pgfpathcurveto{\pgfqpoint{0.466925in}{0.936393in}}{\pgfqpoint{0.487843in}{0.904629in}}{\pgfqpoint{0.515939in}{0.880016in}}%
\pgfpathcurveto{\pgfqpoint{0.544035in}{0.855402in}}{\pgfqpoint{0.578275in}{0.838845in}}{\pgfqpoint{0.615014in}{0.832105in}}%
\pgfpathcurveto{\pgfqpoint{0.651753in}{0.825366in}}{\pgfqpoint{0.689641in}{0.828692in}}{\pgfqpoint{0.724644in}{0.841730in}}%
\pgfpathcurveto{\pgfqpoint{0.759647in}{0.854769in}}{\pgfqpoint{0.790477in}{0.877039in}}{\pgfqpoint{0.813855in}{0.906172in}}%
\pgfpathcurveto{\pgfqpoint{0.837232in}{0.935304in}}{\pgfqpoint{0.852295in}{0.970227in}}{\pgfqpoint{0.857442in}{1.007223in}}%
\pgfpathlineto{\pgfqpoint{0.754907in}{1.021488in}}%
\pgfpathcurveto{\pgfqpoint{0.752334in}{1.002990in}}{\pgfqpoint{0.744802in}{0.985528in}}{\pgfqpoint{0.733113in}{0.970962in}}%
\pgfpathcurveto{\pgfqpoint{0.721425in}{0.956396in}}{\pgfqpoint{0.706009in}{0.945260in}}{\pgfqpoint{0.688508in}{0.938741in}}%
\pgfpathcurveto{\pgfqpoint{0.671006in}{0.932222in}}{\pgfqpoint{0.652063in}{0.930559in}}{\pgfqpoint{0.633693in}{0.933929in}}%
\pgfpathcurveto{\pgfqpoint{0.615323in}{0.937298in}}{\pgfqpoint{0.598204in}{0.945577in}}{\pgfqpoint{0.584156in}{0.957884in}}%
\pgfpathcurveto{\pgfqpoint{0.570108in}{0.970191in}}{\pgfqpoint{0.559648in}{0.986073in}}{\pgfqpoint{0.553891in}{1.003839in}}%
\pgfpathcurveto{\pgfqpoint{0.548134in}{1.021606in}}{\pgfqpoint{0.547290in}{1.040604in}}{\pgfqpoint{0.551450in}{1.058811in}}%
\pgfpathcurveto{\pgfqpoint{0.555610in}{1.077018in}}{\pgfqpoint{0.564620in}{1.093764in}}{\pgfqpoint{0.577522in}{1.107268in}}%
\pgfpathcurveto{\pgfqpoint{0.590424in}{1.120771in}}{\pgfqpoint{0.606743in}{1.130535in}}{\pgfqpoint{0.624742in}{1.135519in}}%
\pgfpathlineto{\pgfqpoint{0.597111in}{1.235286in}}%
\pgfpathclose%
\pgfusepath{stroke,fill}%
\end{pgfscope}%
\begin{pgfscope}%
\pgfsetbuttcap%
\pgfsetmiterjoin%
\definecolor{currentfill}{rgb}{0.266667,0.666667,0.600000}%
\pgfsetfillcolor{currentfill}%
\pgfsetlinewidth{1.007514pt}%
\definecolor{currentstroke}{rgb}{1.000000,1.000000,1.000000}%
\pgfsetstrokecolor{currentstroke}%
\pgfsetdash{}{0pt}%
\pgfpathmoveto{\pgfqpoint{0.857442in}{1.007223in}}%
\pgfpathcurveto{\pgfqpoint{0.861521in}{1.036540in}}{\pgfqpoint{0.859265in}{1.066393in}}{\pgfqpoint{0.850829in}{1.094764in}}%
\pgfpathcurveto{\pgfqpoint{0.842393in}{1.123135in}}{\pgfqpoint{0.827970in}{1.149370in}}{\pgfqpoint{0.808536in}{1.171695in}}%
\pgfpathcurveto{\pgfqpoint{0.789102in}{1.194020in}}{\pgfqpoint{0.765105in}{1.211920in}}{\pgfqpoint{0.738166in}{1.224185in}}%
\pgfpathcurveto{\pgfqpoint{0.711228in}{1.236450in}}{\pgfqpoint{0.681971in}{1.242797in}}{\pgfqpoint{0.652372in}{1.242797in}}%
\pgfpathlineto{\pgfqpoint{0.652372in}{1.139275in}}%
\pgfpathcurveto{\pgfqpoint{0.667171in}{1.139275in}}{\pgfqpoint{0.681800in}{1.136101in}}{\pgfqpoint{0.695269in}{1.129969in}}%
\pgfpathcurveto{\pgfqpoint{0.708738in}{1.123836in}}{\pgfqpoint{0.720737in}{1.114886in}}{\pgfqpoint{0.730454in}{1.103724in}}%
\pgfpathcurveto{\pgfqpoint{0.740171in}{1.092561in}}{\pgfqpoint{0.747382in}{1.079444in}}{\pgfqpoint{0.751601in}{1.065258in}}%
\pgfpathcurveto{\pgfqpoint{0.755819in}{1.051073in}}{\pgfqpoint{0.756946in}{1.036146in}}{\pgfqpoint{0.754907in}{1.021488in}}%
\pgfpathlineto{\pgfqpoint{0.857442in}{1.007223in}}%
\pgfpathclose%
\pgfusepath{stroke,fill}%
\end{pgfscope}%
\begin{pgfscope}%
\definecolor{textcolor}{rgb}{0.000000,0.000000,0.000000}%
\pgfsetstrokecolor{textcolor}%
\pgfsetfillcolor{textcolor}%
\pgftext[x=0.617413in,y=1.290096in,,]{\color{textcolor}{\rmfamily\fontsize{7.500000}{9.000000}\bfseries\selectfont\catcode`\^=\active\def^{\ifmmode\sp\else\^{}\fi}\catcode`\%=\active\def
\end{pgfscope}%
\begin{pgfscope}%
\definecolor{textcolor}{rgb}{0.000000,0.000000,0.000000}%
\pgfsetstrokecolor{textcolor}%
\pgfsetfillcolor{textcolor}%
\pgftext[x=0.441744in,y=0.791820in,,]{\color{textcolor}{\rmfamily\fontsize{7.500000}{9.000000}\bfseries\selectfont\catcode`\^=\active\def^{\ifmmode\sp\else\^{}\fi}\catcode`\%=\active\def
\end{pgfscope}%
\begin{pgfscope}%
\definecolor{textcolor}{rgb}{0.000000,0.000000,0.000000}%
\pgfsetstrokecolor{textcolor}%
\pgfsetfillcolor{textcolor}%
\pgftext[x=0.920362in,y=1.228824in,,]{\color{textcolor}{\rmfamily\fontsize{7.500000}{9.000000}\bfseries\selectfont\catcode`\^=\active\def^{\ifmmode\sp\else\^{}\fi}\catcode`\%=\active\def
\end{pgfscope}%
\begin{pgfscope}%
\pgfsetbuttcap%
\pgfsetmiterjoin%
\definecolor{currentfill}{rgb}{0.800000,0.400000,0.466667}%
\pgfsetfillcolor{currentfill}%
\pgfsetlinewidth{1.007514pt}%
\definecolor{currentstroke}{rgb}{1.000000,1.000000,1.000000}%
\pgfsetstrokecolor{currentstroke}%
\pgfsetdash{}{0pt}%
\pgfpathmoveto{\pgfqpoint{1.359901in}{1.242797in}}%
\pgfpathcurveto{\pgfqpoint{1.350574in}{1.242797in}}{\pgfqpoint{1.341258in}{1.242167in}}{\pgfqpoint{1.332017in}{1.240911in}}%
\pgfpathcurveto{\pgfqpoint{1.322775in}{1.239655in}}{\pgfqpoint{1.313629in}{1.237776in}}{\pgfqpoint{1.304640in}{1.235286in}}%
\pgfpathlineto{\pgfqpoint{1.332271in}{1.135519in}}%
\pgfpathcurveto{\pgfqpoint{1.336765in}{1.136764in}}{\pgfqpoint{1.341338in}{1.137703in}}{\pgfqpoint{1.345959in}{1.138332in}}%
\pgfpathcurveto{\pgfqpoint{1.350580in}{1.138960in}}{\pgfqpoint{1.355238in}{1.139275in}}{\pgfqpoint{1.359901in}{1.139275in}}%
\pgfpathlineto{\pgfqpoint{1.359901in}{1.242797in}}%
\pgfpathclose%
\pgfusepath{stroke,fill}%
\end{pgfscope}%
\begin{pgfscope}%
\pgfsetbuttcap%
\pgfsetmiterjoin%
\definecolor{currentfill}{rgb}{0.266667,0.466667,0.666667}%
\pgfsetfillcolor{currentfill}%
\pgfsetlinewidth{1.007514pt}%
\definecolor{currentstroke}{rgb}{1.000000,1.000000,1.000000}%
\pgfsetstrokecolor{currentstroke}%
\pgfsetdash{}{0pt}%
\pgfpathmoveto{\pgfqpoint{1.304640in}{1.235286in}}%
\pgfpathcurveto{\pgfqpoint{1.267472in}{1.224993in}}{\pgfqpoint{1.233910in}{1.204516in}}{\pgfqpoint{1.207753in}{1.176175in}}%
\pgfpathcurveto{\pgfqpoint{1.181596in}{1.147834in}}{\pgfqpoint{1.163871in}{1.112740in}}{\pgfqpoint{1.156584in}{1.074868in}}%
\pgfpathcurveto{\pgfqpoint{1.149298in}{1.036996in}}{\pgfqpoint{1.152737in}{0.997830in}}{\pgfqpoint{1.166511in}{0.961807in}}%
\pgfpathcurveto{\pgfqpoint{1.180285in}{0.925784in}}{\pgfqpoint{1.203854in}{0.894315in}}{\pgfqpoint{1.234549in}{0.870966in}}%
\pgfpathcurveto{\pgfqpoint{1.265244in}{0.847616in}}{\pgfqpoint{1.301862in}{0.833301in}}{\pgfqpoint{1.340254in}{0.829641in}}%
\pgfpathcurveto{\pgfqpoint{1.378647in}{0.825982in}}{\pgfqpoint{1.417310in}{0.833121in}}{\pgfqpoint{1.451863in}{0.850251in}}%
\pgfpathcurveto{\pgfqpoint{1.486417in}{0.867381in}}{\pgfqpoint{1.515507in}{0.893830in}}{\pgfqpoint{1.535839in}{0.926602in}}%
\pgfpathcurveto{\pgfqpoint{1.556170in}{0.959375in}}{\pgfqpoint{1.566946in}{0.997185in}}{\pgfqpoint{1.566946in}{1.035752in}}%
\pgfpathlineto{\pgfqpoint{1.463424in}{1.035752in}}%
\pgfpathcurveto{\pgfqpoint{1.463424in}{1.016469in}}{\pgfqpoint{1.458036in}{0.997564in}}{\pgfqpoint{1.447870in}{0.981177in}}%
\pgfpathcurveto{\pgfqpoint{1.437704in}{0.964791in}}{\pgfqpoint{1.423159in}{0.951567in}}{\pgfqpoint{1.405882in}{0.943002in}}%
\pgfpathcurveto{\pgfqpoint{1.388605in}{0.934437in}}{\pgfqpoint{1.369274in}{0.930867in}}{\pgfqpoint{1.350078in}{0.932697in}}%
\pgfpathcurveto{\pgfqpoint{1.330881in}{0.934527in}}{\pgfqpoint{1.312573in}{0.941684in}}{\pgfqpoint{1.297225in}{0.953359in}}%
\pgfpathcurveto{\pgfqpoint{1.281877in}{0.965034in}}{\pgfqpoint{1.270093in}{0.980768in}}{\pgfqpoint{1.263206in}{0.998779in}}%
\pgfpathcurveto{\pgfqpoint{1.256319in}{1.016791in}}{\pgfqpoint{1.254600in}{1.036374in}}{\pgfqpoint{1.258243in}{1.055310in}}%
\pgfpathcurveto{\pgfqpoint{1.261886in}{1.074246in}}{\pgfqpoint{1.270748in}{1.091793in}}{\pgfqpoint{1.283827in}{1.105964in}}%
\pgfpathcurveto{\pgfqpoint{1.296905in}{1.120134in}}{\pgfqpoint{1.313687in}{1.130372in}}{\pgfqpoint{1.332271in}{1.135519in}}%
\pgfpathlineto{\pgfqpoint{1.304640in}{1.235286in}}%
\pgfpathclose%
\pgfusepath{stroke,fill}%
\end{pgfscope}%
\begin{pgfscope}%
\pgfsetbuttcap%
\pgfsetmiterjoin%
\definecolor{currentfill}{rgb}{0.266667,0.666667,0.600000}%
\pgfsetfillcolor{currentfill}%
\pgfsetlinewidth{1.007514pt}%
\definecolor{currentstroke}{rgb}{1.000000,1.000000,1.000000}%
\pgfsetstrokecolor{currentstroke}%
\pgfsetdash{}{0pt}%
\pgfpathmoveto{\pgfqpoint{1.566946in}{1.035752in}}%
\pgfpathcurveto{\pgfqpoint{1.566946in}{1.090643in}}{\pgfqpoint{1.545118in}{1.143342in}}{\pgfqpoint{1.506304in}{1.182155in}}%
\pgfpathcurveto{\pgfqpoint{1.467490in}{1.220969in}}{\pgfqpoint{1.414792in}{1.242797in}}{\pgfqpoint{1.359901in}{1.242797in}}%
\pgfpathlineto{\pgfqpoint{1.359901in}{1.139275in}}%
\pgfpathcurveto{\pgfqpoint{1.387346in}{1.139275in}}{\pgfqpoint{1.413696in}{1.128360in}}{\pgfqpoint{1.433103in}{1.108954in}}%
\pgfpathcurveto{\pgfqpoint{1.452509in}{1.089547in}}{\pgfqpoint{1.463424in}{1.063198in}}{\pgfqpoint{1.463424in}{1.035752in}}%
\pgfpathlineto{\pgfqpoint{1.566946in}{1.035752in}}%
\pgfpathclose%
\pgfusepath{stroke,fill}%
\end{pgfscope}%
\begin{pgfscope}%
\definecolor{textcolor}{rgb}{0.000000,0.000000,0.000000}%
\pgfsetstrokecolor{textcolor}%
\pgfsetfillcolor{textcolor}%
\pgftext[x=1.300432in,y=1.296143in,,]{\color{textcolor}{\rmfamily\fontsize{7.500000}{9.000000}\bfseries\selectfont\catcode`\^=\active\def^{\ifmmode\sp\else\^{}\fi}\catcode`\%=\active\def
\end{pgfscope}%
\begin{pgfscope}%
\definecolor{textcolor}{rgb}{0.000000,0.000000,0.000000}%
\pgfsetstrokecolor{textcolor}%
\pgfsetfillcolor{textcolor}%
\pgftext[x=1.172526in,y=0.786240in,,]{\color{textcolor}{\rmfamily\fontsize{7.500000}{9.000000}\bfseries\selectfont\catcode`\^=\active\def^{\ifmmode\sp\else\^{}\fi}\catcode`\%=\active\def
\end{pgfscope}%
\begin{pgfscope}%
\definecolor{textcolor}{rgb}{0.000000,0.000000,0.000000}%
\pgfsetstrokecolor{textcolor}%
\pgfsetfillcolor{textcolor}%
\pgftext[x=1.619811in,y=1.222333in,,]{\color{textcolor}{\rmfamily\fontsize{7.500000}{9.000000}\bfseries\selectfont\catcode`\^=\active\def^{\ifmmode\sp\else\^{}\fi}\catcode`\%=\active\def
\end{pgfscope}%
\begin{pgfscope}%
\pgfsetbuttcap%
\pgfsetmiterjoin%
\definecolor{currentfill}{rgb}{0.800000,0.400000,0.466667}%
\pgfsetfillcolor{currentfill}%
\pgfsetlinewidth{1.007514pt}%
\definecolor{currentstroke}{rgb}{1.000000,1.000000,1.000000}%
\pgfsetstrokecolor{currentstroke}%
\pgfsetdash{}{0pt}%
\pgfpathmoveto{\pgfqpoint{2.065799in}{1.241516in}}%
\pgfpathcurveto{\pgfqpoint{2.024300in}{1.241516in}}{\pgfqpoint{1.983745in}{1.229040in}}{\pgfqpoint{1.949422in}{1.205714in}}%
\pgfpathcurveto{\pgfqpoint{1.915100in}{1.182388in}}{\pgfqpoint{1.888570in}{1.149274in}}{\pgfqpoint{1.873293in}{1.110690in}}%
\pgfpathcurveto{\pgfqpoint{1.858017in}{1.072105in}}{\pgfqpoint{1.854688in}{1.029805in}}{\pgfqpoint{1.863740in}{0.989306in}}%
\pgfpathcurveto{\pgfqpoint{1.872793in}{0.948807in}}{\pgfqpoint{1.893816in}{0.911949in}}{\pgfqpoint{1.924067in}{0.883542in}}%
\pgfpathlineto{\pgfqpoint{1.994933in}{0.959006in}}%
\pgfpathcurveto{\pgfqpoint{1.979807in}{0.973210in}}{\pgfqpoint{1.969296in}{0.991639in}}{\pgfqpoint{1.964770in}{1.011888in}}%
\pgfpathcurveto{\pgfqpoint{1.960243in}{1.032138in}}{\pgfqpoint{1.961908in}{1.053288in}}{\pgfqpoint{1.969546in}{1.072580in}}%
\pgfpathcurveto{\pgfqpoint{1.977184in}{1.091873in}}{\pgfqpoint{1.990449in}{1.108430in}}{\pgfqpoint{2.007611in}{1.120093in}}%
\pgfpathcurveto{\pgfqpoint{2.024772in}{1.131755in}}{\pgfqpoint{2.045050in}{1.137994in}}{\pgfqpoint{2.065799in}{1.137994in}}%
\pgfpathlineto{\pgfqpoint{2.065799in}{1.241516in}}%
\pgfpathclose%
\pgfusepath{stroke,fill}%
\end{pgfscope}%
\begin{pgfscope}%
\pgfsetbuttcap%
\pgfsetmiterjoin%
\definecolor{currentfill}{rgb}{0.266667,0.466667,0.666667}%
\pgfsetfillcolor{currentfill}%
\pgfsetlinewidth{1.007514pt}%
\definecolor{currentstroke}{rgb}{1.000000,1.000000,1.000000}%
\pgfsetstrokecolor{currentstroke}%
\pgfsetdash{}{0pt}%
\pgfpathmoveto{\pgfqpoint{1.924067in}{0.883542in}}%
\pgfpathcurveto{\pgfqpoint{1.950853in}{0.858388in}}{\pgfqpoint{1.983859in}{0.840812in}}{\pgfqpoint{2.019682in}{0.832627in}}%
\pgfpathcurveto{\pgfqpoint{2.055504in}{0.824443in}}{\pgfqpoint{2.092868in}{0.825940in}}{\pgfqpoint{2.127921in}{0.836965in}}%
\pgfpathcurveto{\pgfqpoint{2.162973in}{0.847990in}}{\pgfqpoint{2.194467in}{0.868151in}}{\pgfqpoint{2.219155in}{0.895368in}}%
\pgfpathcurveto{\pgfqpoint{2.243842in}{0.922585in}}{\pgfqpoint{2.260845in}{0.955890in}}{\pgfqpoint{2.268409in}{0.991848in}}%
\pgfpathlineto{\pgfqpoint{2.167104in}{1.013160in}}%
\pgfpathcurveto{\pgfqpoint{2.163322in}{0.995180in}}{\pgfqpoint{2.154821in}{0.978528in}}{\pgfqpoint{2.142477in}{0.964920in}}%
\pgfpathcurveto{\pgfqpoint{2.130133in}{0.951311in}}{\pgfqpoint{2.114386in}{0.941231in}}{\pgfqpoint{2.096860in}{0.935718in}}%
\pgfpathcurveto{\pgfqpoint{2.079334in}{0.930206in}}{\pgfqpoint{2.060652in}{0.929457in}}{\pgfqpoint{2.042740in}{0.933549in}}%
\pgfpathcurveto{\pgfqpoint{2.024829in}{0.937642in}}{\pgfqpoint{2.008326in}{0.946429in}}{\pgfqpoint{1.994933in}{0.959006in}}%
\pgfpathlineto{\pgfqpoint{1.924067in}{0.883542in}}%
\pgfpathclose%
\pgfusepath{stroke,fill}%
\end{pgfscope}%
\begin{pgfscope}%
\pgfsetbuttcap%
\pgfsetmiterjoin%
\definecolor{currentfill}{rgb}{0.266667,0.666667,0.600000}%
\pgfsetfillcolor{currentfill}%
\pgfsetlinewidth{1.007514pt}%
\definecolor{currentstroke}{rgb}{1.000000,1.000000,1.000000}%
\pgfsetstrokecolor{currentstroke}%
\pgfsetdash{}{0pt}%
\pgfpathmoveto{\pgfqpoint{2.268409in}{0.991848in}}%
\pgfpathcurveto{\pgfqpoint{2.274751in}{1.021994in}}{\pgfqpoint{2.274286in}{1.053178in}}{\pgfqpoint{2.267047in}{1.083121in}}%
\pgfpathcurveto{\pgfqpoint{2.259809in}{1.113064in}}{\pgfqpoint{2.245978in}{1.141017in}}{\pgfqpoint{2.226567in}{1.164937in}}%
\pgfpathcurveto{\pgfqpoint{2.207155in}{1.188857in}}{\pgfqpoint{2.182648in}{1.208146in}}{\pgfqpoint{2.154836in}{1.221394in}}%
\pgfpathcurveto{\pgfqpoint{2.127025in}{1.234641in}}{\pgfqpoint{2.096604in}{1.241516in}}{\pgfqpoint{2.065799in}{1.241516in}}%
\pgfpathlineto{\pgfqpoint{2.065799in}{1.137994in}}%
\pgfpathcurveto{\pgfqpoint{2.081202in}{1.137994in}}{\pgfqpoint{2.096412in}{1.134556in}}{\pgfqpoint{2.110318in}{1.127932in}}%
\pgfpathcurveto{\pgfqpoint{2.124223in}{1.121309in}}{\pgfqpoint{2.136477in}{1.111664in}}{\pgfqpoint{2.146183in}{1.099704in}}%
\pgfpathcurveto{\pgfqpoint{2.155889in}{1.087744in}}{\pgfqpoint{2.162804in}{1.073768in}}{\pgfqpoint{2.166423in}{1.058796in}}%
\pgfpathcurveto{\pgfqpoint{2.170042in}{1.043825in}}{\pgfqpoint{2.170275in}{1.028232in}}{\pgfqpoint{2.167104in}{1.013160in}}%
\pgfpathlineto{\pgfqpoint{2.268409in}{0.991848in}}%
\pgfpathclose%
\pgfusepath{stroke,fill}%
\end{pgfscope}%
\begin{pgfscope}%
\definecolor{textcolor}{rgb}{0.000000,0.000000,0.000000}%
\pgfsetstrokecolor{textcolor}%
\pgfsetfillcolor{textcolor}%
\pgftext[x=1.734484in,y=1.071808in,,]{\color{textcolor}{\rmfamily\fontsize{7.500000}{9.000000}\bfseries\selectfont\catcode`\^=\active\def^{\ifmmode\sp\else\^{}\fi}\catcode`\%=\active\def
\end{pgfscope}%
\begin{pgfscope}%
\definecolor{textcolor}{rgb}{0.000000,0.000000,0.000000}%
\pgfsetstrokecolor{textcolor}%
\pgfsetfillcolor{textcolor}%
\pgftext[x=2.223931in,y=0.789541in,,]{\color{textcolor}{\rmfamily\fontsize{7.500000}{9.000000}\bfseries\selectfont\catcode`\^=\active\def^{\ifmmode\sp\else\^{}\fi}\catcode`\%=\active\def
\end{pgfscope}%
\begin{pgfscope}%
\definecolor{textcolor}{rgb}{0.000000,0.000000,0.000000}%
\pgfsetstrokecolor{textcolor}%
\pgfsetfillcolor{textcolor}%
\pgftext[x=2.316665in,y=1.236240in,,]{\color{textcolor}{\rmfamily\fontsize{7.500000}{9.000000}\bfseries\selectfont\catcode`\^=\active\def^{\ifmmode\sp\else\^{}\fi}\catcode`\%=\active\def
\end{pgfscope}%
\begin{pgfscope}%
\pgfsetbuttcap%
\pgfsetmiterjoin%
\definecolor{currentfill}{rgb}{0.800000,0.400000,0.466667}%
\pgfsetfillcolor{currentfill}%
\pgfsetlinewidth{1.007514pt}%
\definecolor{currentstroke}{rgb}{1.000000,1.000000,1.000000}%
\pgfsetstrokecolor{currentstroke}%
\pgfsetdash{}{0pt}%
\pgfpathmoveto{\pgfqpoint{2.773214in}{1.241516in}}%
\pgfpathcurveto{\pgfqpoint{2.761499in}{1.241516in}}{\pgfqpoint{2.749805in}{1.240522in}}{\pgfqpoint{2.738258in}{1.238544in}}%
\pgfpathcurveto{\pgfqpoint{2.726711in}{1.236566in}}{\pgfqpoint{2.715353in}{1.233612in}}{\pgfqpoint{2.704305in}{1.229713in}}%
\pgfpathlineto{\pgfqpoint{2.738760in}{1.132092in}}%
\pgfpathcurveto{\pgfqpoint{2.744283in}{1.134041in}}{\pgfqpoint{2.749962in}{1.135519in}}{\pgfqpoint{2.755736in}{1.136508in}}%
\pgfpathcurveto{\pgfqpoint{2.761509in}{1.137497in}}{\pgfqpoint{2.767356in}{1.137994in}}{\pgfqpoint{2.773214in}{1.137994in}}%
\pgfpathlineto{\pgfqpoint{2.773214in}{1.241516in}}%
\pgfpathclose%
\pgfusepath{stroke,fill}%
\end{pgfscope}%
\begin{pgfscope}%
\pgfsetbuttcap%
\pgfsetmiterjoin%
\definecolor{currentfill}{rgb}{0.266667,0.466667,0.666667}%
\pgfsetfillcolor{currentfill}%
\pgfsetlinewidth{1.007514pt}%
\definecolor{currentstroke}{rgb}{1.000000,1.000000,1.000000}%
\pgfsetstrokecolor{currentstroke}%
\pgfsetdash{}{0pt}%
\pgfpathmoveto{\pgfqpoint{2.704305in}{1.229713in}}%
\pgfpathcurveto{\pgfqpoint{2.673029in}{1.218674in}}{\pgfqpoint{2.644859in}{1.200280in}}{\pgfqpoint{2.622173in}{1.176085in}}%
\pgfpathcurveto{\pgfqpoint{2.599488in}{1.151889in}}{\pgfqpoint{2.582945in}{1.122593in}}{\pgfqpoint{2.573942in}{1.090672in}}%
\pgfpathcurveto{\pgfqpoint{2.564940in}{1.058750in}}{\pgfqpoint{2.563737in}{1.025127in}}{\pgfqpoint{2.570438in}{0.992644in}}%
\pgfpathcurveto{\pgfqpoint{2.577138in}{0.960161in}}{\pgfqpoint{2.591547in}{0.929759in}}{\pgfqpoint{2.612446in}{0.904005in}}%
\pgfpathcurveto{\pgfqpoint{2.633346in}{0.878251in}}{\pgfqpoint{2.660131in}{0.857893in}}{\pgfqpoint{2.690539in}{0.844649in}}%
\pgfpathcurveto{\pgfqpoint{2.720947in}{0.831405in}}{\pgfqpoint{2.754097in}{0.825660in}}{\pgfqpoint{2.787188in}{0.827898in}}%
\pgfpathcurveto{\pgfqpoint{2.820279in}{0.830137in}}{\pgfqpoint{2.852353in}{0.840294in}}{\pgfqpoint{2.880701in}{0.857513in}}%
\pgfpathcurveto{\pgfqpoint{2.909048in}{0.874731in}}{\pgfqpoint{2.932847in}{0.898512in}}{\pgfqpoint{2.950088in}{0.926845in}}%
\pgfpathlineto{\pgfqpoint{2.861651in}{0.980658in}}%
\pgfpathcurveto{\pgfqpoint{2.853031in}{0.966491in}}{\pgfqpoint{2.841131in}{0.954601in}}{\pgfqpoint{2.826957in}{0.945992in}}%
\pgfpathcurveto{\pgfqpoint{2.812784in}{0.937383in}}{\pgfqpoint{2.796747in}{0.932304in}}{\pgfqpoint{2.780201in}{0.931185in}}%
\pgfpathcurveto{\pgfqpoint{2.763655in}{0.930065in}}{\pgfqpoint{2.747080in}{0.932938in}}{\pgfqpoint{2.731876in}{0.939560in}}%
\pgfpathcurveto{\pgfqpoint{2.716672in}{0.946182in}}{\pgfqpoint{2.703280in}{0.956361in}}{\pgfqpoint{2.692830in}{0.969238in}}%
\pgfpathcurveto{\pgfqpoint{2.682380in}{0.982115in}}{\pgfqpoint{2.675176in}{0.997316in}}{\pgfqpoint{2.671826in}{1.013558in}}%
\pgfpathcurveto{\pgfqpoint{2.668476in}{1.029799in}}{\pgfqpoint{2.669077in}{1.046611in}}{\pgfqpoint{2.673578in}{1.062571in}}%
\pgfpathcurveto{\pgfqpoint{2.678080in}{1.078532in}}{\pgfqpoint{2.686351in}{1.093180in}}{\pgfqpoint{2.697694in}{1.105278in}}%
\pgfpathcurveto{\pgfqpoint{2.709036in}{1.117376in}}{\pgfqpoint{2.723122in}{1.126573in}}{\pgfqpoint{2.738760in}{1.132092in}}%
\pgfpathlineto{\pgfqpoint{2.704305in}{1.229713in}}%
\pgfpathclose%
\pgfusepath{stroke,fill}%
\end{pgfscope}%
\begin{pgfscope}%
\pgfsetbuttcap%
\pgfsetmiterjoin%
\definecolor{currentfill}{rgb}{0.266667,0.666667,0.600000}%
\pgfsetfillcolor{currentfill}%
\pgfsetlinewidth{1.007514pt}%
\definecolor{currentstroke}{rgb}{1.000000,1.000000,1.000000}%
\pgfsetstrokecolor{currentstroke}%
\pgfsetdash{}{0pt}%
\pgfpathmoveto{\pgfqpoint{2.950088in}{0.926845in}}%
\pgfpathcurveto{\pgfqpoint{2.969189in}{0.958236in}}{\pgfqpoint{2.979593in}{0.994154in}}{\pgfqpoint{2.980228in}{1.030894in}}%
\pgfpathcurveto{\pgfqpoint{2.980863in}{1.067634in}}{\pgfqpoint{2.971706in}{1.103889in}}{\pgfqpoint{2.953701in}{1.135921in}}%
\pgfpathcurveto{\pgfqpoint{2.935696in}{1.167954in}}{\pgfqpoint{2.909484in}{1.194622in}}{\pgfqpoint{2.877768in}{1.213178in}}%
\pgfpathcurveto{\pgfqpoint{2.846051in}{1.231734in}}{\pgfqpoint{2.809960in}{1.241516in}}{\pgfqpoint{2.773214in}{1.241516in}}%
\pgfpathlineto{\pgfqpoint{2.773214in}{1.137994in}}%
\pgfpathcurveto{\pgfqpoint{2.791587in}{1.137994in}}{\pgfqpoint{2.809633in}{1.133103in}}{\pgfqpoint{2.825491in}{1.123825in}}%
\pgfpathcurveto{\pgfqpoint{2.841349in}{1.114547in}}{\pgfqpoint{2.854455in}{1.101212in}}{\pgfqpoint{2.863457in}{1.085196in}}%
\pgfpathcurveto{\pgfqpoint{2.872460in}{1.069180in}}{\pgfqpoint{2.877039in}{1.051053in}}{\pgfqpoint{2.876721in}{1.032682in}}%
\pgfpathcurveto{\pgfqpoint{2.876404in}{1.014312in}}{\pgfqpoint{2.871201in}{0.996354in}}{\pgfqpoint{2.861651in}{0.980658in}}%
\pgfpathlineto{\pgfqpoint{2.950088in}{0.926845in}}%
\pgfpathclose%
\pgfusepath{stroke,fill}%
\end{pgfscope}%
\begin{pgfscope}%
\definecolor{textcolor}{rgb}{0.000000,0.000000,0.000000}%
\pgfsetstrokecolor{textcolor}%
\pgfsetfillcolor{textcolor}%
\pgftext[x=2.731281in,y=1.288755in,,]{\color{textcolor}{\rmfamily\fontsize{7.500000}{9.000000}\bfseries\selectfont\catcode`\^=\active\def^{\ifmmode\sp\else\^{}\fi}\catcode`\%=\active\def
\end{pgfscope}%
\begin{pgfscope}%
\definecolor{textcolor}{rgb}{0.000000,0.000000,0.000000}%
\pgfsetstrokecolor{textcolor}%
\pgfsetfillcolor{textcolor}%
\pgftext[x=2.468147in,y=0.927183in,,]{\color{textcolor}{\rmfamily\fontsize{7.500000}{9.000000}\bfseries\selectfont\catcode`\^=\active\def^{\ifmmode\sp\else\^{}\fi}\catcode`\%=\active\def
\end{pgfscope}%
\begin{pgfscope}%
\definecolor{textcolor}{rgb}{0.000000,0.000000,0.000000}%
\pgfsetstrokecolor{textcolor}%
\pgfsetfillcolor{textcolor}%
\pgftext[x=3.064010in,y=1.196072in,,]{\color{textcolor}{\rmfamily\fontsize{7.500000}{9.000000}\bfseries\selectfont\catcode`\^=\active\def^{\ifmmode\sp\else\^{}\fi}\catcode`\%=\active\def
\end{pgfscope}%
\begin{pgfscope}%
\pgfsetbuttcap%
\pgfsetmiterjoin%
\definecolor{currentfill}{rgb}{0.800000,0.400000,0.466667}%
\pgfsetfillcolor{currentfill}%
\pgfsetlinewidth{1.007514pt}%
\definecolor{currentstroke}{rgb}{1.000000,1.000000,1.000000}%
\pgfsetstrokecolor{currentstroke}%
\pgfsetdash{}{0pt}%
\pgfpathmoveto{\pgfqpoint{0.652372in}{0.599375in}}%
\pgfpathcurveto{\pgfqpoint{0.602304in}{0.599375in}}{\pgfqpoint{0.553896in}{0.581214in}}{\pgfqpoint{0.516184in}{0.548280in}}%
\pgfpathcurveto{\pgfqpoint{0.478472in}{0.515347in}}{\pgfqpoint{0.453956in}{0.469827in}}{\pgfqpoint{0.447213in}{0.420215in}}%
\pgfpathcurveto{\pgfqpoint{0.440470in}{0.370603in}}{\pgfqpoint{0.451947in}{0.320190in}}{\pgfqpoint{0.479501in}{0.278386in}}%
\pgfpathcurveto{\pgfqpoint{0.507055in}{0.236582in}}{\pgfqpoint{0.548859in}{0.206159in}}{\pgfqpoint{0.597111in}{0.192796in}}%
\pgfpathlineto{\pgfqpoint{0.624742in}{0.292563in}}%
\pgfpathcurveto{\pgfqpoint{0.600616in}{0.299245in}}{\pgfqpoint{0.579714in}{0.314456in}}{\pgfqpoint{0.565936in}{0.335358in}}%
\pgfpathcurveto{\pgfqpoint{0.552159in}{0.356260in}}{\pgfqpoint{0.546421in}{0.381466in}}{\pgfqpoint{0.549793in}{0.406272in}}%
\pgfpathcurveto{\pgfqpoint{0.553164in}{0.431078in}}{\pgfqpoint{0.565422in}{0.453839in}}{\pgfqpoint{0.584278in}{0.470305in}}%
\pgfpathcurveto{\pgfqpoint{0.603134in}{0.486772in}}{\pgfqpoint{0.627338in}{0.495853in}}{\pgfqpoint{0.652372in}{0.495853in}}%
\pgfpathlineto{\pgfqpoint{0.652372in}{0.599375in}}%
\pgfpathclose%
\pgfusepath{stroke,fill}%
\end{pgfscope}%
\begin{pgfscope}%
\pgfsetbuttcap%
\pgfsetmiterjoin%
\definecolor{currentfill}{rgb}{0.266667,0.466667,0.666667}%
\pgfsetfillcolor{currentfill}%
\pgfsetlinewidth{1.007514pt}%
\definecolor{currentstroke}{rgb}{1.000000,1.000000,1.000000}%
\pgfsetstrokecolor{currentstroke}%
\pgfsetdash{}{0pt}%
\pgfpathmoveto{\pgfqpoint{0.597111in}{0.192796in}}%
\pgfpathcurveto{\pgfqpoint{0.627857in}{0.184281in}}{\pgfqpoint{0.660159in}{0.182986in}}{\pgfqpoint{0.691488in}{0.189014in}}%
\pgfpathcurveto{\pgfqpoint{0.722817in}{0.195041in}}{\pgfqpoint{0.752332in}{0.208228in}}{\pgfqpoint{0.777724in}{0.227544in}}%
\pgfpathcurveto{\pgfqpoint{0.803116in}{0.246859in}}{\pgfqpoint{0.823703in}{0.271784in}}{\pgfqpoint{0.837873in}{0.300368in}}%
\pgfpathcurveto{\pgfqpoint{0.852043in}{0.328951in}}{\pgfqpoint{0.859417in}{0.360427in}}{\pgfqpoint{0.859417in}{0.392330in}}%
\pgfpathlineto{\pgfqpoint{0.755895in}{0.392330in}}%
\pgfpathcurveto{\pgfqpoint{0.755895in}{0.376379in}}{\pgfqpoint{0.752208in}{0.360641in}}{\pgfqpoint{0.745122in}{0.346349in}}%
\pgfpathcurveto{\pgfqpoint{0.738037in}{0.332057in}}{\pgfqpoint{0.727744in}{0.319595in}}{\pgfqpoint{0.715048in}{0.309937in}}%
\pgfpathcurveto{\pgfqpoint{0.702352in}{0.300279in}}{\pgfqpoint{0.687594in}{0.293685in}}{\pgfqpoint{0.671930in}{0.290672in}}%
\pgfpathcurveto{\pgfqpoint{0.656266in}{0.287658in}}{\pgfqpoint{0.640115in}{0.288305in}}{\pgfqpoint{0.624742in}{0.292563in}}%
\pgfpathlineto{\pgfqpoint{0.597111in}{0.192796in}}%
\pgfpathclose%
\pgfusepath{stroke,fill}%
\end{pgfscope}%
\begin{pgfscope}%
\pgfsetbuttcap%
\pgfsetmiterjoin%
\definecolor{currentfill}{rgb}{0.266667,0.666667,0.600000}%
\pgfsetfillcolor{currentfill}%
\pgfsetlinewidth{1.007514pt}%
\definecolor{currentstroke}{rgb}{1.000000,1.000000,1.000000}%
\pgfsetstrokecolor{currentstroke}%
\pgfsetdash{}{0pt}%
\pgfpathmoveto{\pgfqpoint{0.859417in}{0.392330in}}%
\pgfpathcurveto{\pgfqpoint{0.859417in}{0.447221in}}{\pgfqpoint{0.837589in}{0.499920in}}{\pgfqpoint{0.798775in}{0.538733in}}%
\pgfpathcurveto{\pgfqpoint{0.759961in}{0.577547in}}{\pgfqpoint{0.707263in}{0.599375in}}{\pgfqpoint{0.652372in}{0.599375in}}%
\pgfpathlineto{\pgfqpoint{0.652372in}{0.495853in}}%
\pgfpathcurveto{\pgfqpoint{0.679817in}{0.495853in}}{\pgfqpoint{0.706167in}{0.484938in}}{\pgfqpoint{0.725574in}{0.465532in}}%
\pgfpathcurveto{\pgfqpoint{0.744980in}{0.446125in}}{\pgfqpoint{0.755895in}{0.419776in}}{\pgfqpoint{0.755895in}{0.392330in}}%
\pgfpathlineto{\pgfqpoint{0.859417in}{0.392330in}}%
\pgfpathclose%
\pgfusepath{stroke,fill}%
\end{pgfscope}%
\begin{pgfscope}%
\definecolor{textcolor}{rgb}{0.000000,0.000000,0.000000}%
\pgfsetstrokecolor{textcolor}%
\pgfsetfillcolor{textcolor}%
\pgftext[x=0.425343in,y=0.619743in,,]{\color{textcolor}{\rmfamily\fontsize{7.500000}{9.000000}\bfseries\selectfont\catcode`\^=\active\def^{\ifmmode\sp\else\^{}\fi}\catcode`\%=\active\def
\end{pgfscope}%
\begin{pgfscope}%
\definecolor{textcolor}{rgb}{0.000000,0.000000,0.000000}%
\pgfsetstrokecolor{textcolor}%
\pgfsetfillcolor{textcolor}%
\pgftext[x=0.808773in,y=0.146699in,,]{\color{textcolor}{\rmfamily\fontsize{7.500000}{9.000000}\bfseries\selectfont\catcode`\^=\active\def^{\ifmmode\sp\else\^{}\fi}\catcode`\%=\active\def
\end{pgfscope}%
\begin{pgfscope}%
\definecolor{textcolor}{rgb}{0.000000,0.000000,0.000000}%
\pgfsetstrokecolor{textcolor}%
\pgfsetfillcolor{textcolor}%
\pgftext[x=0.874027in,y=0.611653in,,]{\color{textcolor}{\rmfamily\fontsize{7.500000}{9.000000}\bfseries\selectfont\catcode`\^=\active\def^{\ifmmode\sp\else\^{}\fi}\catcode`\%=\active\def
\end{pgfscope}%
\begin{pgfscope}%
\pgfsetbuttcap%
\pgfsetmiterjoin%
\definecolor{currentfill}{rgb}{0.800000,0.400000,0.466667}%
\pgfsetfillcolor{currentfill}%
\pgfsetlinewidth{1.007514pt}%
\definecolor{currentstroke}{rgb}{1.000000,1.000000,1.000000}%
\pgfsetstrokecolor{currentstroke}%
\pgfsetdash{}{0pt}%
\pgfpathmoveto{\pgfqpoint{1.359901in}{0.598029in}}%
\pgfpathcurveto{\pgfqpoint{1.323193in}{0.598029in}}{\pgfqpoint{1.287136in}{0.588267in}}{\pgfqpoint{1.255442in}{0.569747in}}%
\pgfpathcurveto{\pgfqpoint{1.223748in}{0.551226in}}{\pgfqpoint{1.197542in}{0.524606in}}{\pgfqpoint{1.179521in}{0.492625in}}%
\pgfpathcurveto{\pgfqpoint{1.161501in}{0.460644in}}{\pgfqpoint{1.152305in}{0.424439in}}{\pgfqpoint{1.152881in}{0.387735in}}%
\pgfpathcurveto{\pgfqpoint{1.153457in}{0.351031in}}{\pgfqpoint{1.163784in}{0.315132in}}{\pgfqpoint{1.182800in}{0.283733in}}%
\pgfpathcurveto{\pgfqpoint{1.201815in}{0.252333in}}{\pgfqpoint{1.228843in}{0.226549in}}{\pgfqpoint{1.261103in}{0.209032in}}%
\pgfpathcurveto{\pgfqpoint{1.293362in}{0.191515in}}{\pgfqpoint{1.329708in}{0.182889in}}{\pgfqpoint{1.366398in}{0.184041in}}%
\pgfpathcurveto{\pgfqpoint{1.403088in}{0.185193in}}{\pgfqpoint{1.438821in}{0.196082in}}{\pgfqpoint{1.469918in}{0.215588in}}%
\pgfpathcurveto{\pgfqpoint{1.501015in}{0.235093in}}{\pgfqpoint{1.526373in}{0.262523in}}{\pgfqpoint{1.543381in}{0.295053in}}%
\pgfpathlineto{\pgfqpoint{1.451641in}{0.343019in}}%
\pgfpathcurveto{\pgfqpoint{1.443137in}{0.326753in}}{\pgfqpoint{1.430458in}{0.313039in}}{\pgfqpoint{1.414910in}{0.303286in}}%
\pgfpathcurveto{\pgfqpoint{1.399361in}{0.293533in}}{\pgfqpoint{1.381495in}{0.288089in}}{\pgfqpoint{1.363149in}{0.287513in}}%
\pgfpathcurveto{\pgfqpoint{1.344804in}{0.286937in}}{\pgfqpoint{1.326632in}{0.291250in}}{\pgfqpoint{1.310502in}{0.300008in}}%
\pgfpathcurveto{\pgfqpoint{1.294372in}{0.308766in}}{\pgfqpoint{1.280858in}{0.321659in}}{\pgfqpoint{1.271350in}{0.337359in}}%
\pgfpathcurveto{\pgfqpoint{1.261843in}{0.353058in}}{\pgfqpoint{1.256679in}{0.371008in}}{\pgfqpoint{1.256391in}{0.389360in}}%
\pgfpathcurveto{\pgfqpoint{1.256103in}{0.407712in}}{\pgfqpoint{1.260701in}{0.425814in}}{\pgfqpoint{1.269711in}{0.441805in}}%
\pgfpathcurveto{\pgfqpoint{1.278722in}{0.457795in}}{\pgfqpoint{1.291824in}{0.471105in}}{\pgfqpoint{1.307671in}{0.480365in}}%
\pgfpathcurveto{\pgfqpoint{1.323518in}{0.489625in}}{\pgfqpoint{1.341547in}{0.494507in}}{\pgfqpoint{1.359901in}{0.494507in}}%
\pgfpathlineto{\pgfqpoint{1.359901in}{0.598029in}}%
\pgfpathclose%
\pgfusepath{stroke,fill}%
\end{pgfscope}%
\begin{pgfscope}%
\pgfsetbuttcap%
\pgfsetmiterjoin%
\definecolor{currentfill}{rgb}{0.266667,0.466667,0.666667}%
\pgfsetfillcolor{currentfill}%
\pgfsetlinewidth{1.007514pt}%
\definecolor{currentstroke}{rgb}{1.000000,1.000000,1.000000}%
\pgfsetstrokecolor{currentstroke}%
\pgfsetdash{}{0pt}%
\pgfpathmoveto{\pgfqpoint{1.543381in}{0.295053in}}%
\pgfpathcurveto{\pgfqpoint{1.564336in}{0.335132in}}{\pgfqpoint{1.571593in}{0.380995in}}{\pgfqpoint{1.564034in}{0.425585in}}%
\pgfpathcurveto{\pgfqpoint{1.556476in}{0.470176in}}{\pgfqpoint{1.534512in}{0.511085in}}{\pgfqpoint{1.501519in}{0.542020in}}%
\pgfpathlineto{\pgfqpoint{1.430710in}{0.466502in}}%
\pgfpathcurveto{\pgfqpoint{1.447206in}{0.451035in}}{\pgfqpoint{1.458189in}{0.430580in}}{\pgfqpoint{1.461968in}{0.408285in}}%
\pgfpathcurveto{\pgfqpoint{1.465747in}{0.385989in}}{\pgfqpoint{1.462119in}{0.363058in}}{\pgfqpoint{1.451641in}{0.343019in}}%
\pgfpathlineto{\pgfqpoint{1.543381in}{0.295053in}}%
\pgfpathclose%
\pgfusepath{stroke,fill}%
\end{pgfscope}%
\begin{pgfscope}%
\pgfsetbuttcap%
\pgfsetmiterjoin%
\definecolor{currentfill}{rgb}{0.266667,0.666667,0.600000}%
\pgfsetfillcolor{currentfill}%
\pgfsetlinewidth{1.007514pt}%
\definecolor{currentstroke}{rgb}{1.000000,1.000000,1.000000}%
\pgfsetstrokecolor{currentstroke}%
\pgfsetdash{}{0pt}%
\pgfpathmoveto{\pgfqpoint{1.501519in}{0.542020in}}%
\pgfpathcurveto{\pgfqpoint{1.482503in}{0.559851in}}{\pgfqpoint{1.460288in}{0.573931in}}{\pgfqpoint{1.436047in}{0.583519in}}%
\pgfpathcurveto{\pgfqpoint{1.411805in}{0.593106in}}{\pgfqpoint{1.385970in}{0.598029in}}{\pgfqpoint{1.359901in}{0.598029in}}%
\pgfpathlineto{\pgfqpoint{1.359901in}{0.494507in}}%
\pgfpathcurveto{\pgfqpoint{1.372935in}{0.494507in}}{\pgfqpoint{1.385853in}{0.492045in}}{\pgfqpoint{1.397974in}{0.487251in}}%
\pgfpathcurveto{\pgfqpoint{1.410095in}{0.482458in}}{\pgfqpoint{1.421202in}{0.475418in}}{\pgfqpoint{1.430710in}{0.466502in}}%
\pgfpathlineto{\pgfqpoint{1.501519in}{0.542020in}}%
\pgfpathclose%
\pgfusepath{stroke,fill}%
\end{pgfscope}%
\begin{pgfscope}%
\definecolor{textcolor}{rgb}{0.000000,0.000000,0.000000}%
\pgfsetstrokecolor{textcolor}%
\pgfsetfillcolor{textcolor}%
\pgftext[x=1.072125in,y=0.248931in,,]{\color{textcolor}{\rmfamily\fontsize{7.500000}{9.000000}\bfseries\selectfont\catcode`\^=\active\def^{\ifmmode\sp\else\^{}\fi}\catcode`\%=\active\def
\end{pgfscope}%
\begin{pgfscope}%
\definecolor{textcolor}{rgb}{0.000000,0.000000,0.000000}%
\pgfsetstrokecolor{textcolor}%
\pgfsetfillcolor{textcolor}%
\pgftext[x=1.692204in,y=0.449050in,,]{\color{textcolor}{\rmfamily\fontsize{7.500000}{9.000000}\bfseries\selectfont\catcode`\^=\active\def^{\ifmmode\sp\else\^{}\fi}\catcode`\%=\active\def
\end{pgfscope}%
\begin{pgfscope}%
\definecolor{textcolor}{rgb}{0.000000,0.000000,0.000000}%
\pgfsetstrokecolor{textcolor}%
\pgfsetfillcolor{textcolor}%
\pgftext[x=1.464209in,y=0.652083in,,]{\color{textcolor}{\rmfamily\fontsize{7.500000}{9.000000}\bfseries\selectfont\catcode`\^=\active\def^{\ifmmode\sp\else\^{}\fi}\catcode`\%=\active\def
\end{pgfscope}%
\begin{pgfscope}%
\pgfsetbuttcap%
\pgfsetmiterjoin%
\definecolor{currentfill}{rgb}{0.800000,0.400000,0.466667}%
\pgfsetfillcolor{currentfill}%
\pgfsetlinewidth{1.007514pt}%
\definecolor{currentstroke}{rgb}{1.000000,1.000000,1.000000}%
\pgfsetstrokecolor{currentstroke}%
\pgfsetdash{}{0pt}%
\pgfpathmoveto{\pgfqpoint{2.065799in}{0.597694in}}%
\pgfpathcurveto{\pgfqpoint{2.025463in}{0.597694in}}{\pgfqpoint{1.985993in}{0.585907in}}{\pgfqpoint{1.952262in}{0.563788in}}%
\pgfpathcurveto{\pgfqpoint{1.918532in}{0.541669in}}{\pgfqpoint{1.891989in}{0.510168in}}{\pgfqpoint{1.875912in}{0.473175in}}%
\pgfpathcurveto{\pgfqpoint{1.859834in}{0.436182in}}{\pgfqpoint{1.854912in}{0.395284in}}{\pgfqpoint{1.861754in}{0.355533in}}%
\pgfpathcurveto{\pgfqpoint{1.868595in}{0.315781in}}{\pgfqpoint{1.886906in}{0.278882in}}{\pgfqpoint{1.914425in}{0.249392in}}%
\pgfpathcurveto{\pgfqpoint{1.941944in}{0.219902in}}{\pgfqpoint{1.977491in}{0.199086in}}{\pgfqpoint{2.016675in}{0.189516in}}%
\pgfpathcurveto{\pgfqpoint{2.055859in}{0.179946in}}{\pgfqpoint{2.096999in}{0.182032in}}{\pgfqpoint{2.135014in}{0.195516in}}%
\pgfpathcurveto{\pgfqpoint{2.173030in}{0.209000in}}{\pgfqpoint{2.206288in}{0.233304in}}{\pgfqpoint{2.230684in}{0.265427in}}%
\pgfpathcurveto{\pgfqpoint{2.255079in}{0.297549in}}{\pgfqpoint{2.269564in}{0.336111in}}{\pgfqpoint{2.272350in}{0.376351in}}%
\pgfpathlineto{\pgfqpoint{2.169074in}{0.383500in}}%
\pgfpathcurveto{\pgfqpoint{2.167682in}{0.363380in}}{\pgfqpoint{2.160439in}{0.344099in}}{\pgfqpoint{2.148241in}{0.328038in}}%
\pgfpathcurveto{\pgfqpoint{2.136044in}{0.311977in}}{\pgfqpoint{2.119414in}{0.299825in}}{\pgfqpoint{2.100407in}{0.293083in}}%
\pgfpathcurveto{\pgfqpoint{2.081399in}{0.286340in}}{\pgfqpoint{2.060829in}{0.285298in}}{\pgfqpoint{2.041237in}{0.290083in}}%
\pgfpathcurveto{\pgfqpoint{2.021645in}{0.294868in}}{\pgfqpoint{2.003872in}{0.305275in}}{\pgfqpoint{1.990112in}{0.320020in}}%
\pgfpathcurveto{\pgfqpoint{1.976352in}{0.334766in}}{\pgfqpoint{1.967197in}{0.353215in}}{\pgfqpoint{1.963776in}{0.373091in}}%
\pgfpathcurveto{\pgfqpoint{1.960356in}{0.392967in}}{\pgfqpoint{1.962817in}{0.413415in}}{\pgfqpoint{1.970855in}{0.431912in}}%
\pgfpathcurveto{\pgfqpoint{1.978894in}{0.450409in}}{\pgfqpoint{1.992165in}{0.466159in}}{\pgfqpoint{2.009031in}{0.477219in}}%
\pgfpathcurveto{\pgfqpoint{2.025896in}{0.488278in}}{\pgfqpoint{2.045631in}{0.494172in}}{\pgfqpoint{2.065799in}{0.494172in}}%
\pgfpathlineto{\pgfqpoint{2.065799in}{0.597694in}}%
\pgfpathclose%
\pgfusepath{stroke,fill}%
\end{pgfscope}%
\begin{pgfscope}%
\pgfsetbuttcap%
\pgfsetmiterjoin%
\definecolor{currentfill}{rgb}{0.266667,0.466667,0.666667}%
\pgfsetfillcolor{currentfill}%
\pgfsetlinewidth{1.007514pt}%
\definecolor{currentstroke}{rgb}{1.000000,1.000000,1.000000}%
\pgfsetstrokecolor{currentstroke}%
\pgfsetdash{}{0pt}%
\pgfpathmoveto{\pgfqpoint{2.272350in}{0.376351in}}%
\pgfpathcurveto{\pgfqpoint{2.275308in}{0.419077in}}{\pgfqpoint{2.264936in}{0.461687in}}{\pgfqpoint{2.242673in}{0.498275in}}%
\pgfpathcurveto{\pgfqpoint{2.220410in}{0.534863in}}{\pgfqpoint{2.187333in}{0.563656in}}{\pgfqpoint{2.148026in}{0.580666in}}%
\pgfpathlineto{\pgfqpoint{2.106913in}{0.485657in}}%
\pgfpathcurveto{\pgfqpoint{2.126566in}{0.477153in}}{\pgfqpoint{2.143104in}{0.462756in}}{\pgfqpoint{2.154236in}{0.444462in}}%
\pgfpathcurveto{\pgfqpoint{2.165368in}{0.426168in}}{\pgfqpoint{2.170553in}{0.404863in}}{\pgfqpoint{2.169074in}{0.383500in}}%
\pgfpathlineto{\pgfqpoint{2.272350in}{0.376351in}}%
\pgfpathclose%
\pgfusepath{stroke,fill}%
\end{pgfscope}%
\begin{pgfscope}%
\pgfsetbuttcap%
\pgfsetmiterjoin%
\definecolor{currentfill}{rgb}{0.266667,0.666667,0.600000}%
\pgfsetfillcolor{currentfill}%
\pgfsetlinewidth{1.007514pt}%
\definecolor{currentstroke}{rgb}{1.000000,1.000000,1.000000}%
\pgfsetstrokecolor{currentstroke}%
\pgfsetdash{}{0pt}%
\pgfpathmoveto{\pgfqpoint{2.148026in}{0.580666in}}%
\pgfpathcurveto{\pgfqpoint{2.135081in}{0.586268in}}{\pgfqpoint{2.121597in}{0.590532in}}{\pgfqpoint{2.107785in}{0.593392in}}%
\pgfpathcurveto{\pgfqpoint{2.093973in}{0.596253in}}{\pgfqpoint{2.079904in}{0.597694in}}{\pgfqpoint{2.065799in}{0.597694in}}%
\pgfpathlineto{\pgfqpoint{2.065799in}{0.494172in}}%
\pgfpathcurveto{\pgfqpoint{2.072852in}{0.494172in}}{\pgfqpoint{2.079886in}{0.493451in}}{\pgfqpoint{2.086792in}{0.492021in}}%
\pgfpathcurveto{\pgfqpoint{2.093698in}{0.490591in}}{\pgfqpoint{2.100440in}{0.488458in}}{\pgfqpoint{2.106913in}{0.485657in}}%
\pgfpathlineto{\pgfqpoint{2.148026in}{0.580666in}}%
\pgfpathclose%
\pgfusepath{stroke,fill}%
\end{pgfscope}%
\begin{pgfscope}%
\definecolor{textcolor}{rgb}{0.000000,0.000000,0.000000}%
\pgfsetstrokecolor{textcolor}%
\pgfsetfillcolor{textcolor}%
\pgftext[x=1.840339in,y=0.180085in,,]{\color{textcolor}{\rmfamily\fontsize{7.500000}{9.000000}\bfseries\selectfont\catcode`\^=\active\def^{\ifmmode\sp\else\^{}\fi}\catcode`\%=\active\def
\end{pgfscope}%
\begin{pgfscope}%
\definecolor{textcolor}{rgb}{0.000000,0.000000,0.000000}%
\pgfsetstrokecolor{textcolor}%
\pgfsetfillcolor{textcolor}%
\pgftext[x=2.351686in,y=0.545211in,,]{\color{textcolor}{\rmfamily\fontsize{7.500000}{9.000000}\bfseries\selectfont\catcode`\^=\active\def^{\ifmmode\sp\else\^{}\fi}\catcode`\%=\active\def
\end{pgfscope}%
\begin{pgfscope}%
\definecolor{textcolor}{rgb}{0.000000,0.000000,0.000000}%
\pgfsetstrokecolor{textcolor}%
\pgfsetfillcolor{textcolor}%
\pgftext[x=2.122074in,y=0.655592in,,]{\color{textcolor}{\rmfamily\fontsize{7.500000}{9.000000}\bfseries\selectfont\catcode`\^=\active\def^{\ifmmode\sp\else\^{}\fi}\catcode`\%=\active\def
\end{pgfscope}%
\begin{pgfscope}%
\pgfsetbuttcap%
\pgfsetmiterjoin%
\definecolor{currentfill}{rgb}{0.800000,0.400000,0.466667}%
\pgfsetfillcolor{currentfill}%
\pgfsetlinewidth{1.007514pt}%
\definecolor{currentstroke}{rgb}{1.000000,1.000000,1.000000}%
\pgfsetstrokecolor{currentstroke}%
\pgfsetdash{}{0pt}%
\pgfpathmoveto{\pgfqpoint{2.773214in}{0.599445in}}%
\pgfpathcurveto{\pgfqpoint{2.724425in}{0.599445in}}{\pgfqpoint{2.677172in}{0.582200in}}{\pgfqpoint{2.639852in}{0.550774in}}%
\pgfpathcurveto{\pgfqpoint{2.602532in}{0.519347in}}{\pgfqpoint{2.577495in}{0.475719in}}{\pgfqpoint{2.569190in}{0.427642in}}%
\pgfpathcurveto{\pgfqpoint{2.560886in}{0.379565in}}{\pgfqpoint{2.569836in}{0.330066in}}{\pgfqpoint{2.594452in}{0.287941in}}%
\pgfpathcurveto{\pgfqpoint{2.619067in}{0.245817in}}{\pgfqpoint{2.657797in}{0.213719in}}{\pgfqpoint{2.703759in}{0.197352in}}%
\pgfpathlineto{\pgfqpoint{2.738487in}{0.294876in}}%
\pgfpathcurveto{\pgfqpoint{2.715505in}{0.303060in}}{\pgfqpoint{2.696140in}{0.319108in}}{\pgfqpoint{2.683833in}{0.340171in}}%
\pgfpathcurveto{\pgfqpoint{2.671525in}{0.361233in}}{\pgfqpoint{2.667050in}{0.385982in}}{\pgfqpoint{2.671202in}{0.410021in}}%
\pgfpathcurveto{\pgfqpoint{2.675354in}{0.434060in}}{\pgfqpoint{2.687873in}{0.455874in}}{\pgfqpoint{2.706533in}{0.471587in}}%
\pgfpathcurveto{\pgfqpoint{2.725193in}{0.487300in}}{\pgfqpoint{2.748819in}{0.495923in}}{\pgfqpoint{2.773214in}{0.495923in}}%
\pgfpathlineto{\pgfqpoint{2.773214in}{0.599445in}}%
\pgfpathclose%
\pgfusepath{stroke,fill}%
\end{pgfscope}%
\begin{pgfscope}%
\pgfsetbuttcap%
\pgfsetmiterjoin%
\definecolor{currentfill}{rgb}{0.266667,0.466667,0.666667}%
\pgfsetfillcolor{currentfill}%
\pgfsetlinewidth{1.007514pt}%
\definecolor{currentstroke}{rgb}{1.000000,1.000000,1.000000}%
\pgfsetstrokecolor{currentstroke}%
\pgfsetdash{}{0pt}%
\pgfpathmoveto{\pgfqpoint{2.703759in}{0.197352in}}%
\pgfpathcurveto{\pgfqpoint{2.741771in}{0.183816in}}{\pgfqpoint{2.782924in}{0.181683in}}{\pgfqpoint{2.822132in}{0.191217in}}%
\pgfpathcurveto{\pgfqpoint{2.861340in}{0.200750in}}{\pgfqpoint{2.896919in}{0.221541in}}{\pgfqpoint{2.924472in}{0.251019in}}%
\pgfpathcurveto{\pgfqpoint{2.952026in}{0.280497in}}{\pgfqpoint{2.970370in}{0.317397in}}{\pgfqpoint{2.977238in}{0.357158in}}%
\pgfpathcurveto{\pgfqpoint{2.984106in}{0.396920in}}{\pgfqpoint{2.979203in}{0.437835in}}{\pgfqpoint{2.963135in}{0.474848in}}%
\pgfpathlineto{\pgfqpoint{2.868174in}{0.433624in}}%
\pgfpathcurveto{\pgfqpoint{2.876209in}{0.415118in}}{\pgfqpoint{2.878660in}{0.394660in}}{\pgfqpoint{2.875226in}{0.374779in}}%
\pgfpathcurveto{\pgfqpoint{2.871792in}{0.354898in}}{\pgfqpoint{2.862620in}{0.336449in}}{\pgfqpoint{2.848843in}{0.321710in}}%
\pgfpathcurveto{\pgfqpoint{2.835066in}{0.306970in}}{\pgfqpoint{2.817277in}{0.296575in}}{\pgfqpoint{2.797673in}{0.291809in}}%
\pgfpathcurveto{\pgfqpoint{2.778069in}{0.287042in}}{\pgfqpoint{2.757493in}{0.288108in}}{\pgfqpoint{2.738487in}{0.294876in}}%
\pgfpathlineto{\pgfqpoint{2.703759in}{0.197352in}}%
\pgfpathclose%
\pgfusepath{stroke,fill}%
\end{pgfscope}%
\begin{pgfscope}%
\pgfsetbuttcap%
\pgfsetmiterjoin%
\definecolor{currentfill}{rgb}{0.266667,0.666667,0.600000}%
\pgfsetfillcolor{currentfill}%
\pgfsetlinewidth{1.007514pt}%
\definecolor{currentstroke}{rgb}{1.000000,1.000000,1.000000}%
\pgfsetstrokecolor{currentstroke}%
\pgfsetdash{}{0pt}%
\pgfpathmoveto{\pgfqpoint{2.963135in}{0.474848in}}%
\pgfpathcurveto{\pgfqpoint{2.947067in}{0.511861in}}{\pgfqpoint{2.920524in}{0.543382in}}{\pgfqpoint{2.886786in}{0.565516in}}%
\pgfpathcurveto{\pgfqpoint{2.853048in}{0.587650in}}{\pgfqpoint{2.813564in}{0.599445in}}{\pgfqpoint{2.773214in}{0.599445in}}%
\pgfpathlineto{\pgfqpoint{2.773214in}{0.495923in}}%
\pgfpathcurveto{\pgfqpoint{2.793389in}{0.495923in}}{\pgfqpoint{2.813131in}{0.490025in}}{\pgfqpoint{2.830000in}{0.478958in}}%
\pgfpathcurveto{\pgfqpoint{2.846869in}{0.467891in}}{\pgfqpoint{2.860140in}{0.452131in}}{\pgfqpoint{2.868174in}{0.433624in}}%
\pgfpathlineto{\pgfqpoint{2.963135in}{0.474848in}}%
\pgfpathclose%
\pgfusepath{stroke,fill}%
\end{pgfscope}%
\begin{pgfscope}%
\definecolor{textcolor}{rgb}{0.000000,0.000000,0.000000}%
\pgfsetstrokecolor{textcolor}%
\pgfsetfillcolor{textcolor}%
\pgftext[x=2.451542in,y=0.362489in,,]{\color{textcolor}{\rmfamily\fontsize{7.500000}{9.000000}\bfseries\selectfont\catcode`\^=\active\def^{\ifmmode\sp\else\^{}\fi}\catcode`\%=\active\def
\end{pgfscope}%
\begin{pgfscope}%
\definecolor{textcolor}{rgb}{0.000000,0.000000,0.000000}%
\pgfsetstrokecolor{textcolor}%
\pgfsetfillcolor{textcolor}%
\pgftext[x=3.005255in,y=0.177244in,,]{\color{textcolor}{\rmfamily\fontsize{7.500000}{9.000000}\bfseries\selectfont\catcode`\^=\active\def^{\ifmmode\sp\else\^{}\fi}\catcode`\%=\active\def
\end{pgfscope}%
\begin{pgfscope}%
\definecolor{textcolor}{rgb}{0.000000,0.000000,0.000000}%
\pgfsetstrokecolor{textcolor}%
\pgfsetfillcolor{textcolor}%
\pgftext[x=2.952820in,y=0.634122in,,]{\color{textcolor}{\rmfamily\fontsize{7.500000}{9.000000}\bfseries\selectfont\catcode`\^=\active\def^{\ifmmode\sp\else\^{}\fi}\catcode`\%=\active\def
\end{pgfscope}%
\begin{pgfscope}%
\definecolor{textcolor}{rgb}{0.000000,0.000000,0.000000}%
\pgfsetstrokecolor{textcolor}%
\pgfsetfillcolor{textcolor}%
\pgftext[x=0.652651in,y=1.401496in,,base]{\color{textcolor}{\rmfamily\fontsize{8.000000}{9.600000}\bfseries\selectfont\catcode`\^=\active\def^{\ifmmode\sp\else\^{}\fi}\catcode`\%=\active\def
\end{pgfscope}%
\begin{pgfscope}%
\definecolor{textcolor}{rgb}{0.000000,0.000000,0.000000}%
\pgfsetstrokecolor{textcolor}%
\pgfsetfillcolor{textcolor}%
\pgftext[x=1.360180in,y=1.401496in,,base]{\color{textcolor}{\rmfamily\fontsize{8.000000}{9.600000}\bfseries\selectfont\catcode`\^=\active\def^{\ifmmode\sp\else\^{}\fi}\catcode`\%=\active\def
\end{pgfscope}%
\begin{pgfscope}%
\definecolor{textcolor}{rgb}{0.000000,0.000000,0.000000}%
\pgfsetstrokecolor{textcolor}%
\pgfsetfillcolor{textcolor}%
\pgftext[x=2.067709in,y=1.401496in,,base]{\color{textcolor}{\rmfamily\fontsize{8.000000}{9.600000}\bfseries\selectfont\catcode`\^=\active\def^{\ifmmode\sp\else\^{}\fi}\catcode`\%=\active\def
\end{pgfscope}%
\begin{pgfscope}%
\definecolor{textcolor}{rgb}{0.000000,0.000000,0.000000}%
\pgfsetstrokecolor{textcolor}%
\pgfsetfillcolor{textcolor}%
\pgftext[x=2.775238in,y=1.401496in,,base]{\color{textcolor}{\rmfamily\fontsize{8.000000}{9.600000}\bfseries\selectfont\catcode`\^=\active\def^{\ifmmode\sp\else\^{}\fi}\catcode`\%=\active\def
\end{pgfscope}%
\begin{pgfscope}%
\definecolor{textcolor}{rgb}{0.000000,0.000000,0.000000}%
\pgfsetstrokecolor{textcolor}%
\pgfsetfillcolor{textcolor}%
\pgftext[x=0.174721in, y=0.892388in, left, base,rotate=90.000000]{\color{textcolor}{\rmfamily\fontsize{8.000000}{9.600000}\bfseries\selectfont\catcode`\^=\active\def^{\ifmmode\sp\else\^{}\fi}\catcode`\%=\active\def
\end{pgfscope}%
\begin{pgfscope}%
\definecolor{textcolor}{rgb}{0.000000,0.000000,0.000000}%
\pgfsetstrokecolor{textcolor}%
\pgfsetfillcolor{textcolor}%
\pgftext[x=0.177472in, y=0.100000in, left, base,rotate=90.000000]{\color{textcolor}{\rmfamily\fontsize{8.000000}{9.600000}\bfseries\selectfont\catcode`\^=\active\def^{\ifmmode\sp\else\^{}\fi}\catcode`\%=\active\def
\end{pgfscope}%
\end{pgfpicture}%
\makeatother%
\endgroup%

%% file: figures/paper_legend.pgf
\begingroup%
\makeatletter%
\begin{pgfpicture}%
\pgfpathrectangle{\pgfpointorigin}{\pgfqpoint{5.579752in}{0.105166in}}%
\pgfusepath{use as bounding box, clip}%
\begin{pgfscope}%
\pgfsetbuttcap%
\pgfsetmiterjoin%
\definecolor{currentfill}{rgb}{1.000000,1.000000,1.000000}%
\pgfsetfillcolor{currentfill}%
\pgfsetlinewidth{0.000000pt}%
\definecolor{currentstroke}{rgb}{1.000000,1.000000,1.000000}%
\pgfsetstrokecolor{currentstroke}%
\pgfsetdash{}{0pt}%
\pgfpathmoveto{\pgfqpoint{-0.000000in}{0.000000in}}%
\pgfpathlineto{\pgfqpoint{5.579752in}{0.000000in}}%
\pgfpathlineto{\pgfqpoint{5.579752in}{0.105166in}}%
\pgfpathlineto{\pgfqpoint{-0.000000in}{0.105166in}}%
\pgfpathlineto{\pgfqpoint{-0.000000in}{0.000000in}}%
\pgfpathclose%
\pgfusepath{fill}%
\end{pgfscope}%
\begin{pgfscope}%
\pgfsetbuttcap%
\pgfsetmiterjoin%
\definecolor{currentfill}{rgb}{0.145098,0.145098,0.145098}%
\pgfsetfillcolor{currentfill}%
\pgfsetlinewidth{0.000000pt}%
\definecolor{currentstroke}{rgb}{0.000000,0.000000,0.000000}%
\pgfsetstrokecolor{currentstroke}%
\pgfsetstrokeopacity{0.000000}%
\pgfsetdash{}{0pt}%
\pgfpathmoveto{\pgfqpoint{0.000000in}{0.020166in}}%
\pgfpathlineto{\pgfqpoint{0.133333in}{0.020166in}}%
\pgfpathlineto{\pgfqpoint{0.133333in}{0.105166in}}%
\pgfpathlineto{\pgfqpoint{0.000000in}{0.105166in}}%
\pgfpathlineto{\pgfqpoint{0.000000in}{0.020166in}}%
\pgfpathclose%
\pgfusepath{fill}%
\end{pgfscope}%
\begin{pgfscope}%
\definecolor{textcolor}{rgb}{0.000000,0.000000,0.000000}%
\pgfsetstrokecolor{textcolor}%
\pgfsetfillcolor{textcolor}%
\pgftext[x=0.177778in,y=0.024055in,left,base]{\color{textcolor}{\rmfamily\fontsize{8.000000}{9.600000}\selectfont\catcode`\^=\active\def^{\ifmmode\sp\else\^{}\fi}\catcode`\%=\active\def
\end{pgfscope}%
\begin{pgfscope}%
\pgfsetbuttcap%
\pgfsetmiterjoin%
\definecolor{currentfill}{rgb}{0.321569,0.321569,0.321569}%
\pgfsetfillcolor{currentfill}%
\pgfsetlinewidth{0.000000pt}%
\definecolor{currentstroke}{rgb}{0.000000,0.000000,0.000000}%
\pgfsetstrokecolor{currentstroke}%
\pgfsetstrokeopacity{0.000000}%
\pgfsetdash{}{0pt}%
\pgfpathmoveto{\pgfqpoint{0.921663in}{0.020166in}}%
\pgfpathlineto{\pgfqpoint{1.054997in}{0.020166in}}%
\pgfpathlineto{\pgfqpoint{1.054997in}{0.105166in}}%
\pgfpathlineto{\pgfqpoint{0.921663in}{0.105166in}}%
\pgfpathlineto{\pgfqpoint{0.921663in}{0.020166in}}%
\pgfpathclose%
\pgfusepath{fill}%
\end{pgfscope}%
\begin{pgfscope}%
\definecolor{textcolor}{rgb}{0.000000,0.000000,0.000000}%
\pgfsetstrokecolor{textcolor}%
\pgfsetfillcolor{textcolor}%
\pgftext[x=1.099441in,y=0.024055in,left,base]{\color{textcolor}{\rmfamily\fontsize{8.000000}{9.600000}\selectfont\catcode`\^=\active\def^{\ifmmode\sp\else\^{}\fi}\catcode`\%=\active\def
\end{pgfscope}%
\begin{pgfscope}%
\pgfsetbuttcap%
\pgfsetmiterjoin%
\definecolor{currentfill}{rgb}{0.588235,0.588235,0.588235}%
\pgfsetfillcolor{currentfill}%
\pgfsetlinewidth{0.000000pt}%
\definecolor{currentstroke}{rgb}{0.000000,0.000000,0.000000}%
\pgfsetstrokecolor{currentstroke}%
\pgfsetstrokeopacity{0.000000}%
\pgfsetdash{}{0pt}%
\pgfpathmoveto{\pgfqpoint{1.799994in}{0.020166in}}%
\pgfpathlineto{\pgfqpoint{1.933328in}{0.020166in}}%
\pgfpathlineto{\pgfqpoint{1.933328in}{0.105166in}}%
\pgfpathlineto{\pgfqpoint{1.799994in}{0.105166in}}%
\pgfpathlineto{\pgfqpoint{1.799994in}{0.020166in}}%
\pgfpathclose%
\pgfusepath{fill}%
\end{pgfscope}%
\begin{pgfscope}%
\definecolor{textcolor}{rgb}{0.000000,0.000000,0.000000}%
\pgfsetstrokecolor{textcolor}%
\pgfsetfillcolor{textcolor}%
\pgftext[x=1.977772in,y=0.024055in,left,base]{\color{textcolor}{\rmfamily\fontsize{8.000000}{9.600000}\selectfont\catcode`\^=\active\def^{\ifmmode\sp\else\^{}\fi}\catcode`\%=\active\def
\end{pgfscope}%
\begin{pgfscope}%
\pgfsetbuttcap%
\pgfsetmiterjoin%
\definecolor{currentfill}{rgb}{0.741176,0.741176,0.741176}%
\pgfsetfillcolor{currentfill}%
\pgfsetlinewidth{0.000000pt}%
\definecolor{currentstroke}{rgb}{0.000000,0.000000,0.000000}%
\pgfsetstrokecolor{currentstroke}%
\pgfsetstrokeopacity{0.000000}%
\pgfsetdash{}{0pt}%
\pgfpathmoveto{\pgfqpoint{2.929545in}{0.020166in}}%
\pgfpathlineto{\pgfqpoint{3.062878in}{0.020166in}}%
\pgfpathlineto{\pgfqpoint{3.062878in}{0.105166in}}%
\pgfpathlineto{\pgfqpoint{2.929545in}{0.105166in}}%
\pgfpathlineto{\pgfqpoint{2.929545in}{0.020166in}}%
\pgfpathclose%
\pgfusepath{fill}%
\end{pgfscope}%
\begin{pgfscope}%
\definecolor{textcolor}{rgb}{0.000000,0.000000,0.000000}%
\pgfsetstrokecolor{textcolor}%
\pgfsetfillcolor{textcolor}%
\pgftext[x=3.107323in,y=0.024055in,left,base]{\color{textcolor}{\rmfamily\fontsize{8.000000}{9.600000}\selectfont\catcode`\^=\active\def^{\ifmmode\sp\else\^{}\fi}\catcode`\%=\active\def
\end{pgfscope}%
\begin{pgfscope}%
\pgfsetbuttcap%
\pgfsetmiterjoin%
\definecolor{currentfill}{rgb}{0.850980,0.850980,0.850980}%
\pgfsetfillcolor{currentfill}%
\pgfsetlinewidth{0.000000pt}%
\definecolor{currentstroke}{rgb}{0.000000,0.000000,0.000000}%
\pgfsetstrokecolor{currentstroke}%
\pgfsetstrokeopacity{0.000000}%
\pgfsetdash{}{0pt}%
\pgfpathmoveto{\pgfqpoint{4.036207in}{0.020166in}}%
\pgfpathlineto{\pgfqpoint{4.169541in}{0.020166in}}%
\pgfpathlineto{\pgfqpoint{4.169541in}{0.105166in}}%
\pgfpathlineto{\pgfqpoint{4.036207in}{0.105166in}}%
\pgfpathlineto{\pgfqpoint{4.036207in}{0.020166in}}%
\pgfpathclose%
\pgfusepath{fill}%
\end{pgfscope}%
\begin{pgfscope}%
\definecolor{textcolor}{rgb}{0.000000,0.000000,0.000000}%
\pgfsetstrokecolor{textcolor}%
\pgfsetfillcolor{textcolor}%
\pgftext[x=4.213985in,y=0.024055in,left,base]{\color{textcolor}{\rmfamily\fontsize{8.000000}{9.600000}\selectfont\catcode`\^=\active\def^{\ifmmode\sp\else\^{}\fi}\catcode`\%=\active\def
\end{pgfscope}%
\begin{pgfscope}%
\pgfsetbuttcap%
\pgfsetmiterjoin%
\definecolor{currentfill}{rgb}{0.266667,0.666667,0.600000}%
\pgfsetfillcolor{currentfill}%
\pgfsetlinewidth{0.000000pt}%
\definecolor{currentstroke}{rgb}{0.000000,0.000000,0.000000}%
\pgfsetstrokecolor{currentstroke}%
\pgfsetstrokeopacity{0.000000}%
\pgfsetdash{}{0pt}%
\pgfpathmoveto{\pgfqpoint{5.185977in}{0.020166in}}%
\pgfpathlineto{\pgfqpoint{5.319310in}{0.020166in}}%
\pgfpathlineto{\pgfqpoint{5.319310in}{0.105166in}}%
\pgfpathlineto{\pgfqpoint{5.185977in}{0.105166in}}%
\pgfpathlineto{\pgfqpoint{5.185977in}{0.020166in}}%
\pgfpathclose%
\pgfusepath{fill}%
\end{pgfscope}%
\begin{pgfscope}%
\definecolor{textcolor}{rgb}{0.000000,0.000000,0.000000}%
\pgfsetstrokecolor{textcolor}%
\pgfsetfillcolor{textcolor}%
\pgftext[x=5.363754in,y=0.024055in,left,base]{\color{textcolor}{\rmfamily\fontsize{8.000000}{9.600000}\selectfont\catcode`\^=\active\def^{\ifmmode\sp\else\^{}\fi}\catcode`\%=\active\def
\end{pgfscope}%
\end{pgfpicture}%
\makeatother%
\endgroup%

%% file: conclusion.tex
\section{Conclusion}
This work presented a carbon-aware routing framework for function-calling LLMs across heterogeneous edge-cloud tiers. A lightweight k-NN predictor estimates per-query success, delay, and power, guiding the router toward the lowest-emission tier meeting accuracy requirements.
Experiments on BFCL V2 and GeoEngine benchmarks and four LLM families show that our approach maintains competitive accuracy and inference delay close to cloud levels, while reducing emissions by $4\times$ on average and up to $8\times$ for simpler queries, demonstrating that query-level heterogeneity, when properly exploited, becomes a practical path to sustainable agentic AI.

%% file: acknowledgement.tex
\section*{Acknowledgments}
This work is supported by grant NSF 2324854. Any opinions, findings, and conclusions or recommendations expressed in this material are those of the authors and do not necessarily reflect the views of the National Science Foundation.